\pdfoutput=1
\documentclass[review]{elsarticle}
\usepackage{subfigure}
\usepackage{longtable}
\usepackage{algorithm, algorithmic}
\usepackage{bm}
\usepackage{booktabs}
\usepackage{multirow}
\usepackage{geometry}
\usepackage{amsfonts,amssymb}
\usepackage{amsmath}
\usepackage{array}
\usepackage{wrapfig}
\usepackage{graphicx}
\usepackage{hyperref}

\usepackage{lineno,hyperref}
\modulolinenumbers[5]

\journal{arXiv}

\begin{document}

\begin{frontmatter}

\title{Online Continual Learning via the Knowledge Invariant and Spread-out Properties}


\author[mymainaddress]{Ya-nan Han}

\author[mymainaddress]{Jian-wei Liu\corref{mycorrespondingauthor}}
\cortext[mycorrespondingauthor]{Corresponding author}
\ead{liujw@cup.edu.cn}

\address[mymainaddress]{Department of Automation, College of Information Science and Engineering, China University of Petroleum Beijing, Beijing, China
}

\begin{abstract}
The goal of continual learning is to provide intelligent agents that are capable of learning continually a sequence of tasks using the knowledge obtained from previous tasks while performing well on prior tasks. However, a key challenge in this continual learning paradigm is catastrophic forgetting, namely adapting a model to new tasks often leads to severe performance degradation on prior tasks. Current memory-based approaches show their success in alleviating the catastrophic forgetting problem by replaying examples from past tasks when a new task is learned. However, these methods are infeasible to transfer the structural knowledge from previous tasks i.e., similarities or dissimilarities between different instances. Furthermore, the learning bias between the current task and prior tasks is also an urgent problem that should be solved. In this work, we propose a new method, named Online Continual Learning via the Knowledge Invariant and Spread-out Properties (OCLKISP), in which we constraint the evolution of the embedding features via Knowledge Invariant and Spread-out Properties (KISP). Thus, we can further transfer the inter-instance structural knowledge of previous tasks while alleviating the forgetting due to the learning bias. We empirically evaluate our proposed method on four popular benchmarks for continual learning: Split CIFAR 100, Split SVHN, Split CUB200, and Split Tiny-Image-Net. The experimental results show the efficacy of our proposed method compared to the state-of-the-art continual learning algorithms.
\end{abstract}

\begin{keyword}
continual learning; catastrophic forgetting; knowledge invariant and spread-out properties; knowledge transfer.
\end{keyword}

\end{frontmatter}

\section{Introduction}

With the development of deep learning, current deep models can perform well on various computer vision problems\cite{1Girshick2015ICCV,2krizhevsky2012imagenet,3long2015fully,4zhang2018fully}, such as object detection
\cite{5alom2021inception}, image classification
\cite{2krizhevsky2012imagenet,6chandra2020ae}, semantic segmentation
\cite{7pereira2019adaptive}, and so on. The standard learning process of current deep models is underpinned by the assumption that the data is independently and identically distributed (i.i.d). However, in our ever-changing world, the training data often be obtained from an open environment
\cite{8zhou2021co}, which is with stream format
\cite{9golab2003data}. To solve this challenge, the model should learn new tasks incrementally instead of restarting the training process from scratch. A naïve method is to finetune the model on the incoming new tasks, however, it suffers the forgetting of past tasks when new ones are trained, a phenomenon known as catastrophic forgetting
\cite{10robins1993catastrophic}: due to the absence of previous data, the performance on prior tasks drops significantly. Continual learning
\cite{11ring1995continual,12thrun1998learning,13thrun1994lifelong} aims to learn a sequence of tasks by leveraging the knowledge obtained in the past while not forgetting how to perform previous tasks. Fig.1 illustrates the setup of continual learning. As is shown in Fig.1, when the first task arrives, the model is required to classify dogs and spiders. After that, the model is incrementally updated with another new task, i.e. , bag and box, and it needs to correctly classify both the past task (dogs and spiders) and the current new task (bag and box), simultaneously. That means, when the new task arrives sequentially, the agent is required to learn a new task while should not forget how to address the previous tasks.

\begin{figure}[H]
	\centering
	\includegraphics[width=\textwidth]{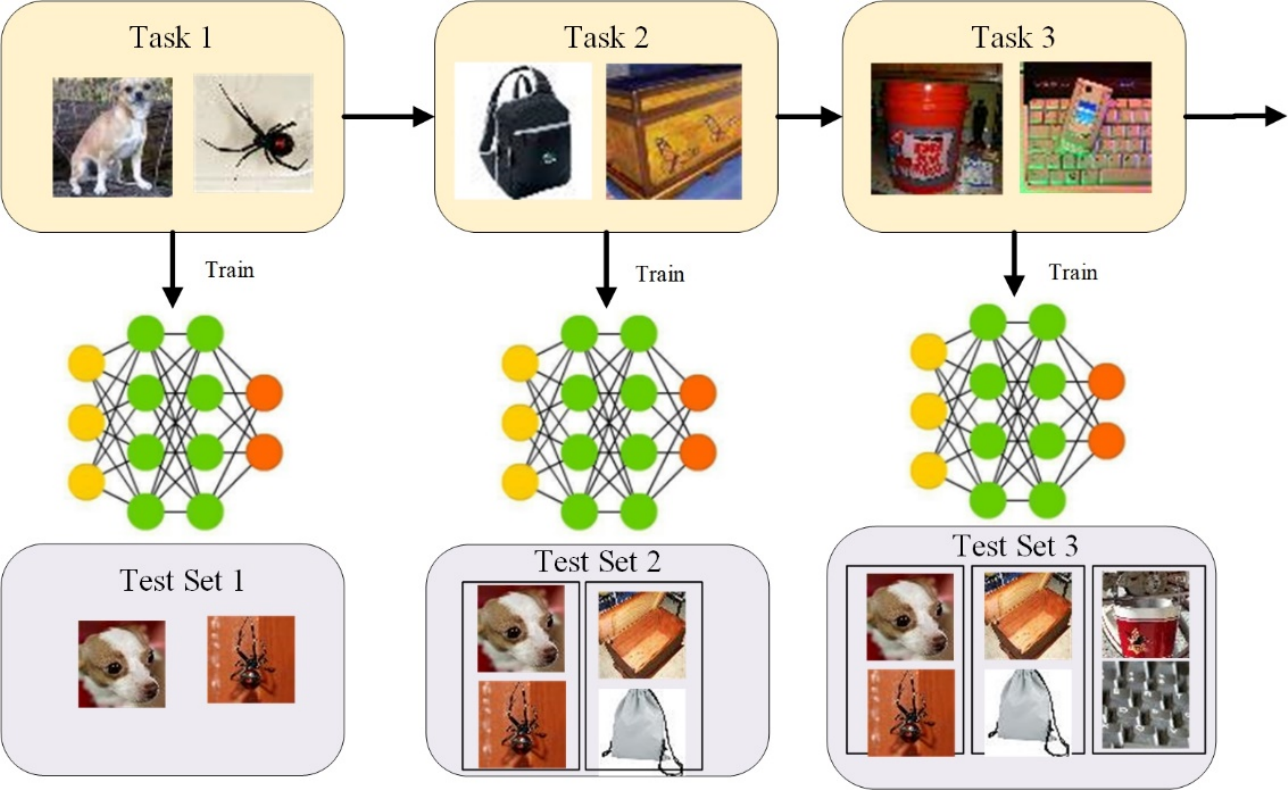}
	\caption{Paradigm for continual learning. When the task comes sequentially, we need to learn a classifier for all tasks incrementally. However, the protocol only allows us to access to the data of current tasks. After completing the learning stage of every task, we test the performance of the current model among all seen tasks, i.e., it should perform well on the current task, while effectively preventing catastrophic forgetting.}
	\label{fig1}
\end{figure}

Recently, several works have attempted to resolve catastrophic forgetting for a continual learning scene, which can be coarsely grouped into two categories according to whether maintain a memory. The methods without memory
\cite{14li2016learning,15zenke2017continual,16kirkpatrick2017overcoming,17chaudhry2018riemannian} constrain the updating of the model parameters to resist catastrophic forgetting and the key difference for this category of approach lies in the way of parameter importance measurement, such as knowledge distillation loss 
\cite{14li2016learning}, the Fisher information matrix
\cite{16kirkpatrick}, and so on. Other works selectively save some examples to prevent forgetting
\cite{18lopez2017gradient,19pham2020bilevel,20chaudhry2019continual,21riemer2018learning}. Compared to other methods, the approaches based on episodic memory can effectively prevent forgetting and achieve superior performance. Thus, in this paper, we also maintain an episodic memory which is used to jointly train to prevent forgetting. However, in this continual learning setting, it is worthy to note that since the budget for storing the examples from the previous tasks is limited for practical continual learning, which can create a bias towards the current task. Such bias makes the model more prone to the current tasks by which worsens forgetting. Recalling the existing researches on this issue, some approaches
\cite{22castro2018end,23wu2019large} attempt to apply an additional subset of examples to alleviate this problem, however, this kind of method requires the whole data of current task which is unavailable for the online continual learning setup. Furthermore, the knowledge distillation strategy used in LWF
\cite{14li2016learning} is to minimize the KL divergence between the soft-max scores (input called logit) of the old and the new tasks. Others
\cite{24hou2019learning,25ni2021alleviate} distill knowledge from the penultimate layer. Zhou et.al
\cite{26zhou2019m2kd} directly leverage all previous model snapshots which means that directly matches the outputs of current network with those from the corresponding previous task models. However, we find that they treat all logits of the intermediate feature levels as separation. That means, these methods only require the current model to mimic the previous model’s logit output values, thus neglecting the important structural knowledge existing in the previous tasks, i.e., similarities or dissimilarities between the logits. We anticipate that fully transferring the important intrinsic structural knowledge underlying in the old tasks is benefit to current task learning, thus further resisting forgetting. In addition, in recent studies, existing works mainly focus on the situations where the whole task’s training data arrives at each step and allow the agent to learn the current tasks for many epochs. However, this assumption is not suitable for many real-world scenarios where the training data is observed sequentially in a stream form. Based on this fact, in this paper, we consider a more realistic but difficult scenario, called online continual learning where the agent only has one look at each instance. It is obvious that continual learning is closer to animal learning\cite{wixted2004psychology}.

In response to the above problems, motivated by positive concentrated and negative separated in the self-supervised field
\cite{27ye2019unsupervised}, we propose a new framework under the online continual learning setting, called Online Continual Learning via the Knowledge Invariant and Spread-out Properties (OCLKISP). Particularly, we propose a new method to transfer the important structural knowledge of the previous tasks via knowledge invariant and spread-out properties for the continual learning setting. In detail, facing the incoming of new tasks, the model should not forget previous knowledge learned from past tasks. That means, the inter-instance space location relationships should keep invariant, in other words, the embedding features of episodic memory learned from the previous model and current model respectively should maintain invariant or the highest similarity as much as possible. However, the embedding features of different examples are separated, which can be beneficial to learn the similarities or dissimilarities of the output distribution among different examples in episodic memory and then better maintain the structure of the embedding features as the task arrives. Here, we transform the multi-class classification problem to a binary classification problem and apply maximum likelihood estimation (MLE) to optimize the continual learning effect.

In summary, the major contributions of this work are as follows:

\begin{itemize}
\item[$\bullet$]We proposed a novel online continual learning method via the knowledge invariant and spread-out properties (OCLKISP), which can transfer the structural knowledge from previous tasks and then improve the overall performance of the model.
\end{itemize}

\begin{itemize}
\item[$\bullet$]We analyze the learning bias of the replay-based method in online continual learning from the angle of embedding features and the classification accuracy respectively. Then, the forgetting problem due to the learning bias can be alleviated via our OCLKISP method. 
\end{itemize}

\begin{itemize}
\item[$\bullet$]We evaluate our proposed method on four popular datasets, and the experimental results illustrate the effectiveness of our OCLKISP method in a continual learning setting. The ablation studies are conducted to further understand the contribution of KISP.
\end{itemize}

The remaining of this paper is organized as follows. In Section 2, we give a survey of the related works. In Section 3, we illustrate the problem of continual learning. In Section 4, we introduce the proposed approach of this work. In Section 5, contrast experiments and ablation studies are performed to demonstrate the effectiveness of our proposed method. Finally, conclusions and future researches are given in Section 6.

\section{Related Work}

\textbf{1)	Continual learning}

Continual learning is now a popular topic in the field of machine learning
\cite{15zenke2017continual},
\cite{282020Adversarial,29xu2021adaptive,30delange2021continual,31daruna2021continual}. The ultimate goal of continual learning is to achieve good performance on the current task while not forgetting how to perform past tasks. According to whether maintain a memory, the method of continual learning can be coarsely grouped into two groups: Non-memory-based strategy and Memory-based strategy.

\textbf{Non-memory-based strategy}	This line of works avoids preserving raw inputs and alleviating memory requirements. We can further divide these methods into parameter-based, distillation-based, and architecture-based methods. Firstly, the methods of parameter-based such as EWC, SI, et al, attempt to use fixed-capacity models and add a regularization term on the loss function to control the changes on significant parameters. The differences among these works lie in the way to estimate the weight importance. For example, a representative method called Elastic Weight Consolidation(EWC)
\cite{16kirkpatrick2017overcoming} applies the Fisher information matrix to estimate the important parameters and then prevent the large change for the important parameters. In Synaptic Intelligence (SI)
\cite{15zenke2017continual}, the importance of weight is computed by the cumulative change of distance after learning new tasks, where the importance is estimated in an online manner during train. Compared to EWC, SI is more intuitive and persuasive in terms of the measure about the importance. In addition, Memory Aware Synapses (MAS)
\cite{32aljundi2018memory} applies the sensitivity of the learned function instead of loss to measure the importance of weights. Another line of work is distillation-based methods and a typical model is Learning Without Forgetting (LWF)
\cite{14li2016learning}, which constrains the change of the model predictions on past tasks via a distillation loss
\cite{33hinton2015distilling}. Although the methods lessen the catastrophic forgetting, they usually fail to solve the forgetting problem well and perform non-favorably when the sequence of tasks is long
\cite{34tu2020extending}. Others focus on the network structure which tries to train a separate model for each incoming task to resist catastrophic forgetting. For example, Progressive Neural Network (PNN)
\cite{35rusu2016progressive} alleviates catastrophic forgetting by fixing the parameters of the network from past tasks while learning new tasks by expanding the network. Other relevant works
\cite{36yoon2018lifelong,37mallya2018piggyback,38mallya2018packnet,39rebuffi2017icarl} consider adaptively selecting the network structure based on the correlation of knowledge and then improving the efficiency of the network capacity. 

\textbf{Memory-based strategy}	This line of work saves prototype instances of former tasks in episodic memory and then a rehearsal or pseudo-rehearsal process is used to prevent catastrophic forgetting when learning a new task. Incremental Classifier and Representation Learning (ICARL) 
\cite{39rebuffi2017icarl} combines knowledge distillation and feature learning to resist forgetting. Specifically, the examples closest to the feature mean of every class are picked and stored in fixed memory. During training, the distillation loss between targets obtained from previous model predictions and current model predictions is applied to alleviate forgetting. Gradient Episodic Memory (GEM) 
\cite{18lopez2017gradient} and Average-GEM (AGEM)
\cite{40chaudhry2019efficient} apply an inequality constraint via episodic memory to prevent the forgetting for former tasks. Hou et al
\cite{24hou2019learning} propose to learn a unified classifier via a cosine linear layer and then incrementally learn new knowledge in a continual learning setting. Wu et al
\cite{23wu2019large} and Pham et al
\cite{19pham2020bilevel} utilize the samples to build an extra validation set and then further improve the generalization of the model. Along with the development of the generative model, such as auto-encoder
\cite{41bengio2013representation}, VAE
\cite{42kingma2013auto}, or GAN
\cite{43goodfellow2020generative}, another direction is to use a generative model to mimic the samples of former tasks and then we prevent forgetting by replaying the generated data of previous tasks
\cite{23wu2019large,44shin2017continual,45ostapenko2019learning}. Compared to the methods of Non-memory based, the major drawback of this kind of method is to require additional computation and storage of raw input samples.

\textbf{2)	Data imbalance}
	
Recently, the method with experience replay has achieved significant improvements in continual learning setting. However, in a replay approach, the data from prior tasks are limited compared to current task which would cause the data imbalance between the previous tasks and current task and then create a learning bias. Many researchers have attempted to resolve this issue in batch continual learning. Francisco et al.
\cite{22castro2018end} attempt to apply an additional balanced subset of examples to finetuning the model. Yue et al.
\cite{46wu2019large} apply a bias correction layer to further avoid the strong bias towards current task. Bowen et al.
\cite{47zhao2020maintaining} attempt to correct the bias via a weight alignment approach in the fully-connected layer after each update. However, these methods require that the data of current task is available and then train multiple epochs or many learning stages is applied to alleviate the learning bias, which is not appropriate for online continual learning where the data of each task arrives in the data stream. We also note that some distillation-based methods
\cite{19pham2020bilevel,26zhou2019m2kd,33hinton2015distilling} can be used to alleviate this problem, such as requiring the corresponding layer in the previous model to be exactly the same as the current model. However, these methods only require the current model to mimic the previous output value of each layer in episodic memory, which is not beneficial to capture the structure knowledge of previous tasks. Based on the above study, in this work, we design a new loss function to deal with the learning bias.

\textbf{3)	Contrastive Learning}	

Recently, contrastive learning 
\cite{hadsell2006dimensionality,weinberger2009distance,chen2020simple}
 has shown great potential to learn general representations by constructing positive and negative examples. There are some prior works in this area. Typically, a random data augmentation, such as scaling, crop, and flip, is employed to every image and then leverages the augmented image as a positive instance. That means we aim to learn more concentration features that should be invariant under different data augmentations. Then, a small batch of randomly selected examples can be considered as negative examples of each image. Under this assumption, some works 
\cite{chen2020simple,ye2019unsupervised,he2020momentum}
 attempt to separate each instance from other negative instances which leads to a spread-out property. Xiao et al.\cite{xiao2021region} consider local features of training inputs and different inputs as positive and negative examples respectively. Tan et al. \cite{tan2022hyperspherical} proposed hyperspherical consistency regularization (HCR) to regularize the classifier which would enforce data points on hyperspheres to have similar structures.

Contrastive representation learning 
\cite{tian2020contrastive,ye2019unsupervised}
 has also shown superior downstream task performance via the positive concentrated and negative separated properties. In the Multiview setting, Tian et al. \cite{tian2020contrastive} consider that every view is noisy and incomplete, however, the important information should be shared among all views. Based on this assumption, different views of input are recognized as the positive instance and then the model can further capture the shared information. Lin et al.\cite{lin2021self} adopt this property to video representation learning. Contrastive learning is now also widely applied to other computer vision fields, e.g., objective detection \cite{silva2021novelty}, natural language domain \cite{fu2021lrc}, and signature verification \cite{tsourounis2022text}. In this work, inspired by the positive concentrated and negative separated properties, we mainly apply this property to the continual learning setting based on our novel observation.

\section{Problem Formulation}

\textbf{Continual Learning}	Before diving into the details of our approach proposed in this work, we first introduce the definition of continual learning followed by a classical memory-based continual learning scenario. Continual learning studies the problem of learning from the streaming tasks, with the goal of incrementally gaining knowledge from data of tasks. Assume there is a sequence of task $\left\{ {1,2, \cdots ,t, \cdots ,{\rm T}} \right\}$, where ${\rm T}$ denotes the total number of training tasks. Each task $t$ has its own dataset $D^t {\rm{ = }}\left\{ {\left( {x_i^t ,\,y_i^t } \right)} \right\}_{i = 1}^{N_t } $,where $x_i^t $ is the i-th training instance and $y_i^t $ is the corresponding ground truth label. It is worth noting that there is no overlapping class among different tasks. During the training process of task $t$, the previous data $D^1 ,D^2 , \cdots ,D^{{\kern 1pt} t{\rm{ - }}1} $ are no longer available. Our goal at every step is not only to learn the knowledge from the current task, but also to retain the knowledge learned from former tasks. After ending every task, we evaluate the performance of the trained model on all currently seen tasks.

The crux problem is that when a new task $t$ arrives, the model should incrementally obtain the ability to correctly classify the new class and all currently seen classes, simultaneously. Here, we split the trained model into two parts: embedding function $\phi \left( \cdot \right)$ and a linear classification head $W^t $. When the training instance $x_i^t $ is observed, firstly, the embedding feature $f_i^t {\text{ = }}\phi \left( {x_i^t } \right)$ is generated by the embedding function. After that, we compute its classification probability for class $y_i^t $ by $\sigma \left( {\left[ {W^t } \right]^T f_i^t } \right)$ where $\sigma \left(\cdot\right)$ is the soft-max operator. Similarly, the classification probability about the all currently seen class label set is calculated as $\sigma \left( {\left[ W \right]^T f_i^t } \right)$, where $W{\text{ = }}\left[ {W^1 ,W^2 , \cdots ,W^t } \right]$ denotes the concatenation of all the current classification heads, i.e., the set of logits for tasks $1, \cdots ,t$. As aforementioned, normally, memory-based approaches prevent catastrophic forgetting via a tiny episodic memory. A typical way is to compute cross-entropy using these examples from episodic memory and current task respectively:

\begin{equation}
L_{CE} \left( {x^t ,y^t ,M{\kern 1pt} ^t } \right){\text{ =   -  E}}_{\left( {x\,^t ,\;y\,^t } \right) \cup M\,^t } \sum\limits_{c = 1}^C {{\mathbf{1}}_{\left[ {c = y\,^t } \right]} } \sigma \left( {\left[ W \right]^T \phi \left( {x\,^t } \right)} \right)
\label{1}
\end{equation}

where $M{\kern 1pt} ^t$ is episodic memory. We optimize the above cross-entropy loss about all previous replay examples in episodic memory and current novel examples, thus learning new knowledge while alleviating the forgetting.

Eq.(1) describes a way to prevent forgetting for continual learning by memory replay. Recently, a lot of literature has shown that compared with other state-of-the-art methods, methods based on experience replay can achieve competitive performance in the online continual learning setting. However, there are some intractable problems which should be solved for these approaches. Firstly, note that the training model only observes none or a few examples of previous tasks due to the small episodic memory but substantially more for the current task. Under this circumstance, as new tasks are learned sequentially, the training process can create a bias toward the current task, thus leading to the forgetting of knowledge learned from previous tasks.  

Here, to more closely and clearly show what the problem is, we make a quantitative analysis. Specifically, we measure the change in cosine distance for embedding features of episodic memory in the buffer after each update when a new task comes. As is illustrated in Fig. 2, due to unbalance data and learning bias, as new knowledge is continuously learned, the change of the embedding features in the buffer gradually increases. Moreover, in Fig. 3, we plot the evolution of task classification accuracy accordingly. We observe that the task classification accuracy performs poorly as new tasks are learned. 

From Figs. 2 and 3, we found that the knowledge learned from previous tasks changes. Thus, the forgetting uninterruptedly and deterioratingly occurs as new knowledge is learned sequentially. So, we must further take some actions to maintain the knowledge invariant and then prevent forgetting due to the learning bias. 

Fortunately, we find that there is some side information neglected in this setting. Specifically, maintaining the structured knowledge in episodic memory (i.e., similarities or dissimilarities between the embedding features $f_i^{\,t} $ and $f_j^{\,t} $) as the new task sequentially arrives is useful means. By means of preserving the structured knowledge in the continual learning setup it is promising to alleviate the learning bias.

\begin{figure}[H]
	\centering
	\subfigure[On the CIFAR100 dataset]{
		\includegraphics[width=7cm]{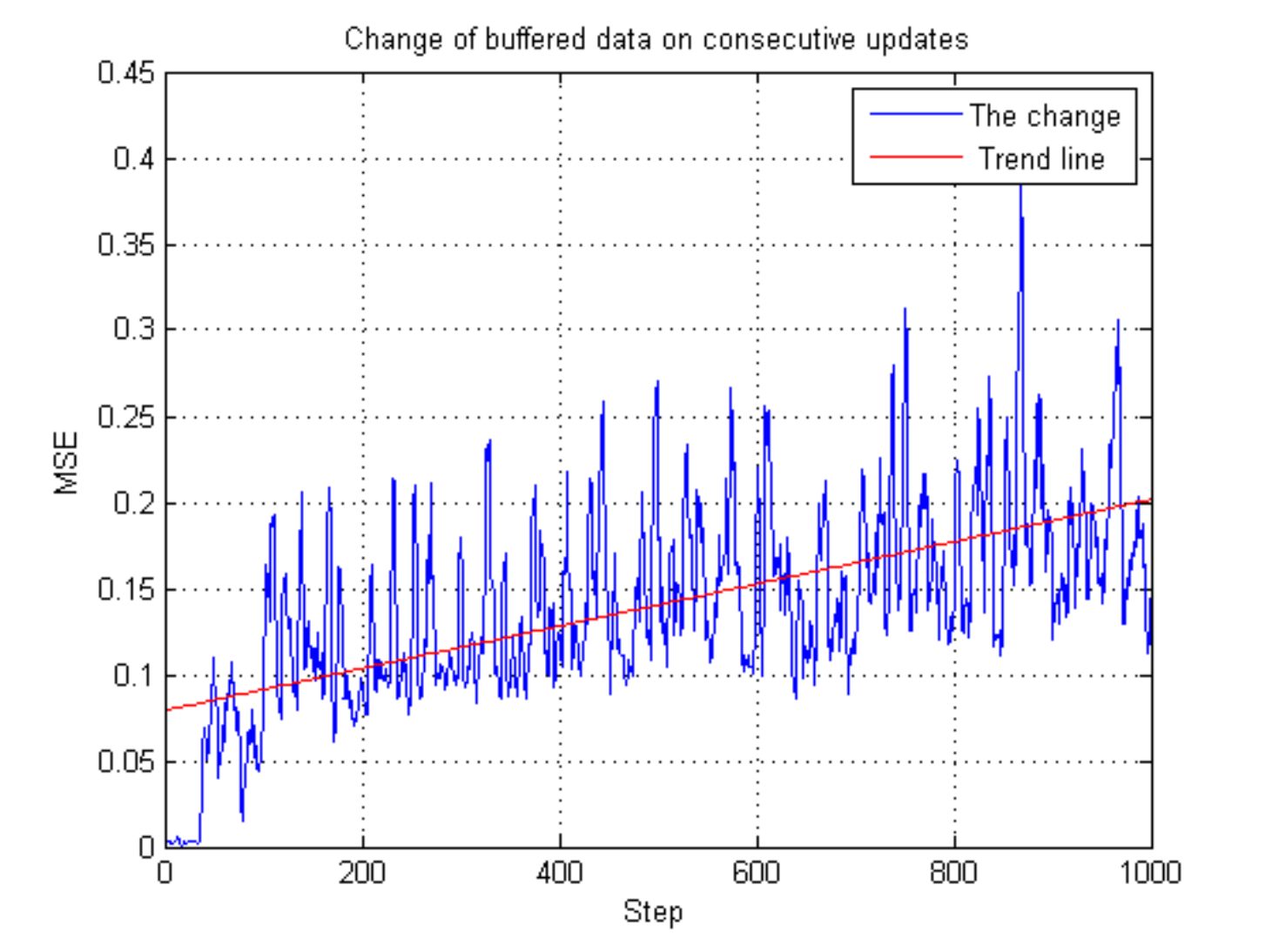}
	}
     \quad
	\subfigure[On the Tiny-Image-Net dataset]{
	\includegraphics[width=7cm]{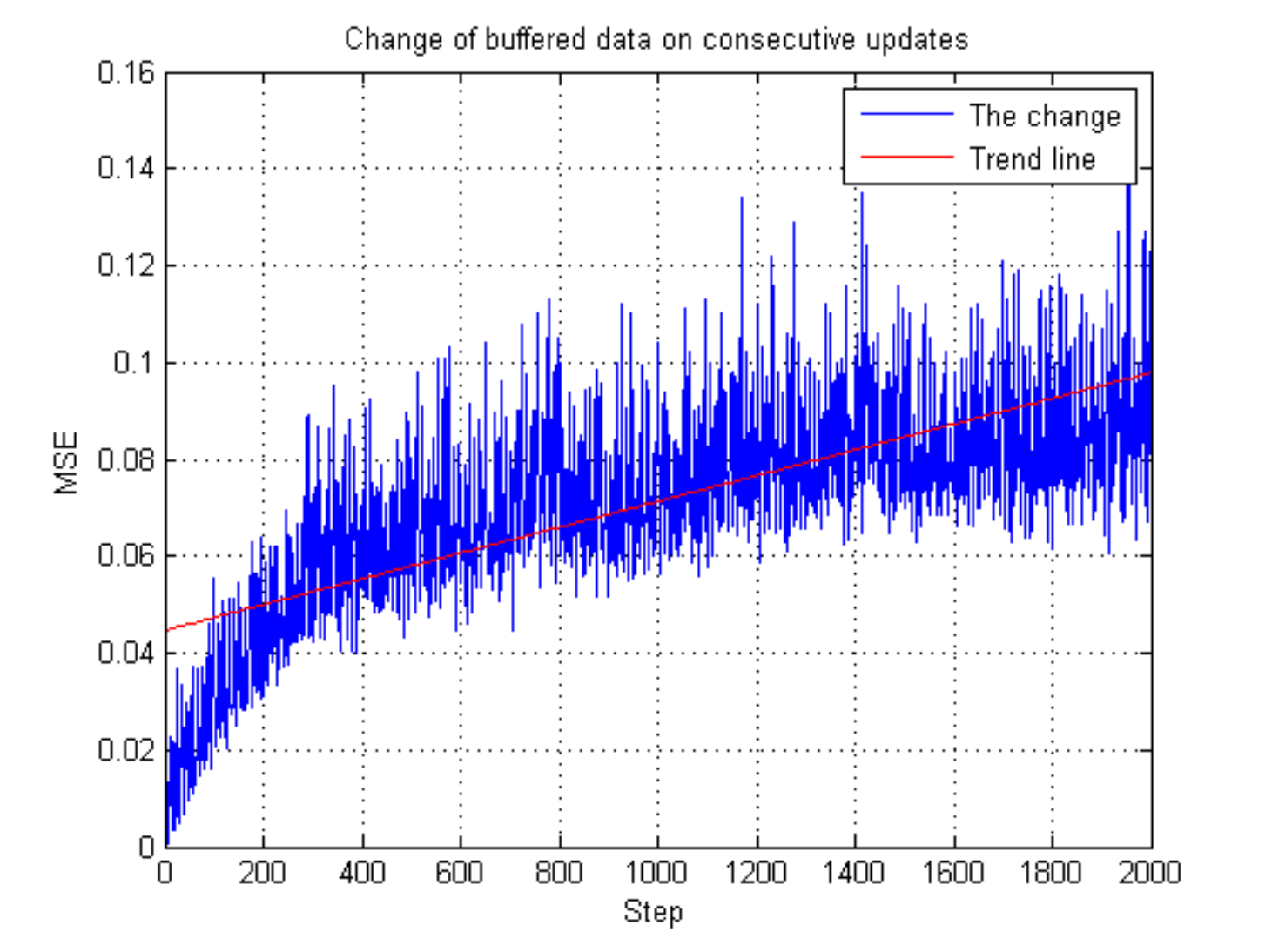}
	}
	\caption{The change curve of embedding features on consecutive tasks on the CIFAR100 and Tiny-Image-Net datasets respectively. The difference is measured by cosines distance in the buffer when new task is learned. We observe that when the new task arrives, the difference increases as new knowledge is learned and the red line indicates the trend of change.}
     \label{fig2}
\end{figure}

\begin{figure}[H]
	\centering
	\subfigure[On the CIFAR100 dataset]{
		\includegraphics[width=7cm]{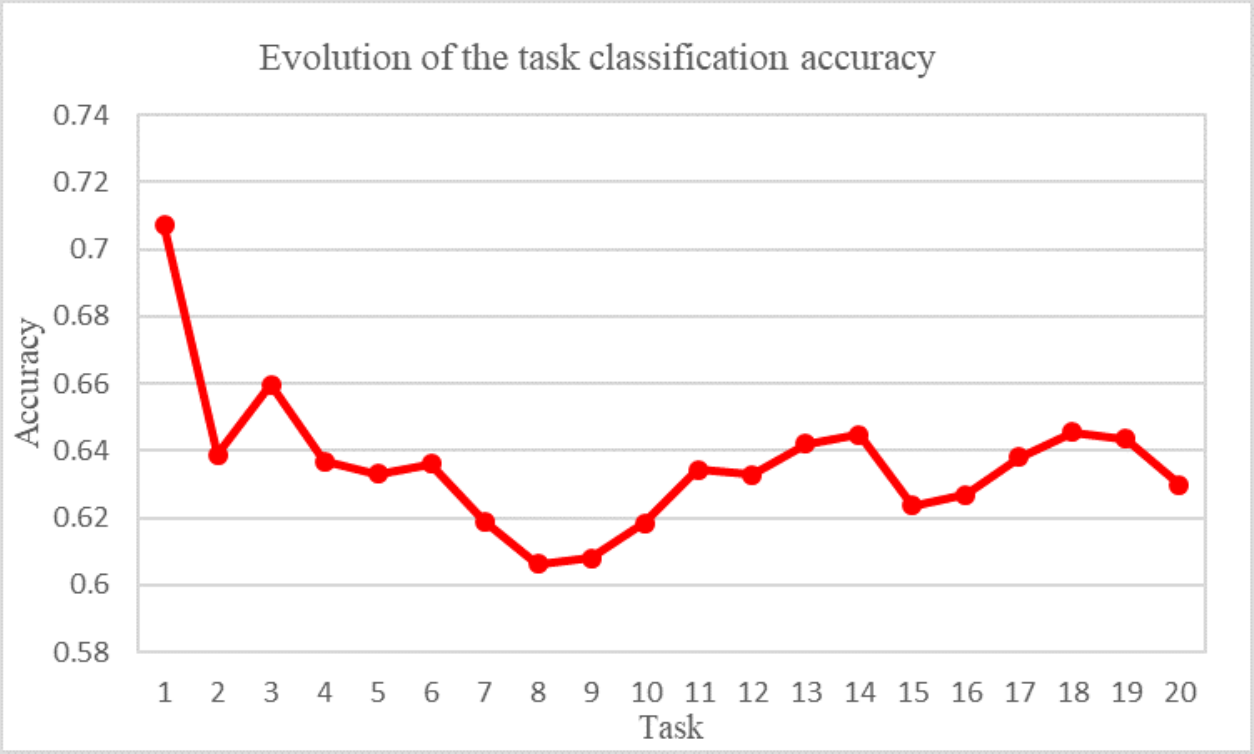}
	}
     \quad
	\subfigure[On the Tiny-Image-Net dataset]{
	\includegraphics[width=7cm]{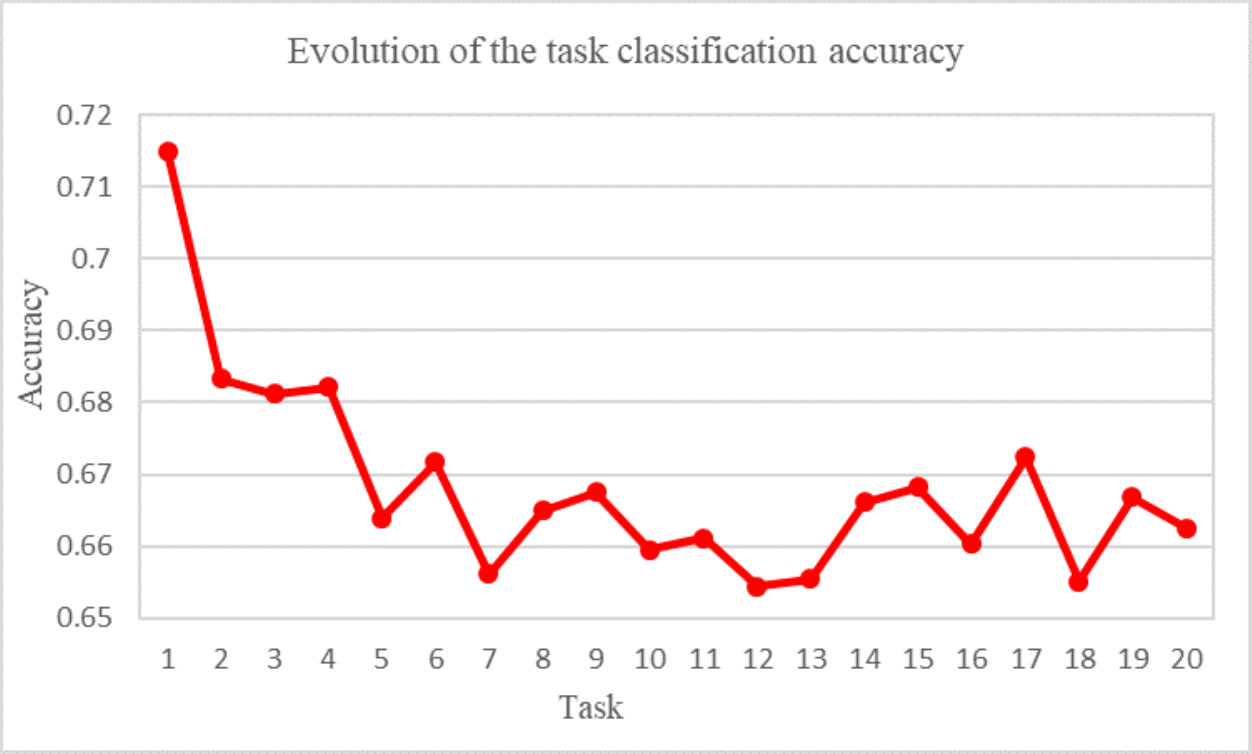}
	}
	\caption{Evolution curves of the task classification accuracies when new tasks come on the CIFAR100 and Tiny-Image-Net datasets respectively.}
     \label{fig3}
\end{figure}

\section{Online Continual Learning via the Knowledge Invariant and Spread-out Properties (OCLKISP)}

In this section, we introduce our proposed OCLKISP for online continual learning detailly. In subsection 4.1, we start the approach discussion with a brief introduction to the framework of OCLKISP. Then, we present the details of our proposed OCLKISP in subsection 4.2. Lastly, we further make the rationale analysis for our proposed approach in subsection 4.3.

\subsection{The Proposed OCLKISP Method}

Different from the typic memory-based method, in our model we further apply the knowledge invariant and spread-out properties to capture the important structural knowledge of the previous tasks, i.e., similarities or dissimilarities between the embedding features $f_i^t $ and $f_j^t $. Specifically, we aim to maintain the following properties: i) \emph{knowledge invariant}, the embedding features of episodic memory learned from the previous tasks and current task respectively should keep invariant or the highest similarity as much as possible, which can encourage the model to maintain the prior knowledge learned from the old tasks; ii) \emph{knowledge separated,} the embedding features of different examples are separated, which can be beneficial to learn the similarities or dissimilarities of the embedding features in episodic memory and then better maintain the structure of the embedding features as the task arrives. In this way, our OCLKISP model is better able to resist forgetting and improve it’s the performance of the model. A visual explanation is shown in Fig. 4.

\begin{figure}[H]
	\centering
	\includegraphics[width=\textwidth]{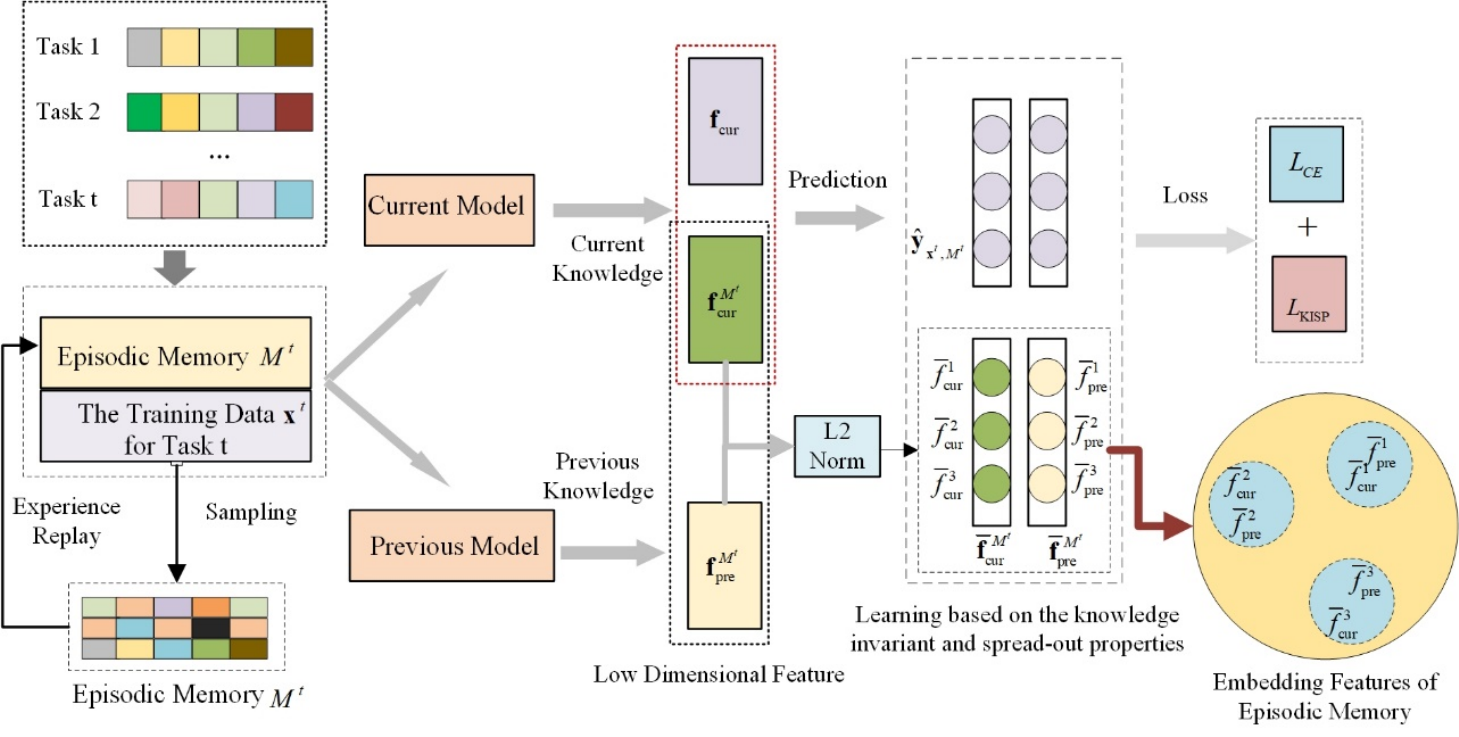}
	\caption{ Illustration of the flowchart of our proposed OCLKISP approach.}
	\label{fig4}
\end{figure}

In Fig. 4, we use $M{\kern 1pt} ^t $ and ${\mathbf{x}}{\kern 1pt} ^t $ to represent episodic memory and the current training data, respectively. ${\mathbf{f}}{\kern 1pt} _{{\text{cur}}} $ and ${\mathbf{f}}{\kern 1pt} _{{\text{cur}}}^{M^t } $ represent the low dimensional embedding features of the current training data and episodic memory from the current model, respectively. ${\mathbf{f}}{\kern 1pt} _{{\text{pre}}}^{M^t } $ represents the low dimensional embedding features of episodic memory from the previous models. The input low dimensional features ${\mathbf{f}}{\kern 1pt} _{{\text{cur}}}^{M^t } $ and ${\mathbf{f}}{\kern 1pt} _{{\text{pre}}}^{M^t } $ are normalized via $l_2 $ are normalized via the L2 norm to acquire the embedding features. After that, the model builds the classifier from these mixed training data where the episodic memory is used to prevent forgetting. We apply the knowledge invariant and spread-out properties to regulate the evolving process of representation. Finally, $L_{CE} $ and $L_{{\text{KISP}}} $ represent cross-entropy loss and constraint item based on knowledge invariant and spread-out properties, respectively.

\subsection{OCLKISP}

The OCLKISP approach proposed in this paper is described in detail below. Specifically, at phase t, that means when the task $t$ comes, we perform memory relay via a tiny episodic memory $M{\kern 1pt} ^t {\text{ = }}\left\{ {x_1 ,x_2 , \cdots ,x_m } \right\}$ where $m$ denotes the total number of episodic memory at phase t. Then, we record the output of episodic memory as  ${\mathbf{f}}_{{\text{cur}}}^{M^t } {\text{ = }}\phi \left( {M^t } \right)$ in the penultimate layer. To measure different features more efficiently and fairly, we map the low dimensional features into the normalized embedding features ${\mathbf{\bar f}}_{{\text{cur}}}^{M^t } $, where ${\mathbf{\bar f}}_{{\text{cur}}}^{M^t } {\text{ = }}{{{\mathbf{f}}_{{\text{cur}}}^{M^t } } \mathord{\left/
 {\vphantom {{{\mathbf{f}}_{{\text{cur}}}^{M^t } } {\left\| {{\mathbf{f}}_{{\text{cur}}}^{M^t } } \right\|}}} \right.
 \kern-\nulldelimiterspace} {\left\| {{\mathbf{f}}_{{\text{cur}}}^{M^t } } \right\|}}$ denotes the $L_2$ normalized. We use ${\mathbf{\bar f}}_{{\text{pre}}}^{M^t }$ to represent the normalized embedding features of the episodic memory from the previous model. 

To further transfer the structural knowledge of episodic memory and prevent forgetting, in this work, we attempt to directly solve it as a binary classification problem via maximum likelihood estimation (MLE). In particular, for instance $x_i  \in M^t $, the embedding feature $\bar f_{{\text{pre}}}^i $ and $\bar f_{{\text{cur}}}^i $ from the current model and previous model respectively should be classified into label $i$, and other instance $x_j  \in M^t ,i \ne j$ should not be categorized into label $i$. By this way, we can effectively preserve the geometric configuration of the episodic memory to resist forgetting. Thus, the probability of the current embedding feature $\bar f_{{\text{cur}}}^i $ being recognized as the label $i$ is denoted by

\begin{equation}
p\left( {i\left| {\bar f_{{\text{cur}}}^i } \right.} \right) = \frac{{\exp \left( {{{\left( {\bar f_{{\text{pre}}}^i } \right)^T \bar f_{{\text{cur}}}^i } \mathord{\left/
 {\vphantom {{\left( {\bar f_{{\text{pre}}}^i } \right)^T \bar f_{{\text{cur}}}^i } \tau }} \right.
 \kern-\nulldelimiterspace} \tau }} \right)}}
{{\sum\nolimits_{k = 1}^m {\exp \left( {{{\left( {\bar f_{{\text{pre}}}^k } \right)^T \bar f_{{\text{cur}}}^i } \mathord{\left/
 {\vphantom {{\left( {\bar f_{{\text{pre}}}^k } \right)^T \bar f_{{\text{cur}}}^i } \tau }} \right.
 \kern-\nulldelimiterspace} \tau }} \right)} }}
\label{2}
\end{equation}

Here,$\tau $ is a scaling factor. Thus, the probability of $\bar f_{{\text{cur}}}^j $ being considered as the label $i$ is given by, 

\begin{equation}
p\left( {i\left| {\bar f_{{\text{cur}}}^j } \right.} \right) = \frac{{\exp \left( {{{\left( {\bar f_{{\text{pre}}}^i } \right)^T \bar f_{{\text{cur}}}^j } \mathord{\left/
 {\vphantom {{\left( {\bar f_{{\text{pre}}}^i } \right)^T \bar f_{{\text{cur}}}^j } \tau }} \right.
 \kern-\nulldelimiterspace} \tau }} \right)}}
{{\sum\nolimits_{k = 1}^m {\exp \left( {{{\left( {\bar f_{{\text{pre}}}^k } \right)^T \bar f_{{\text{cur}}}^j } \mathord{\left/
 {\vphantom {{\left( {\bar f_{{\text{pre}}}^k } \right)^T \bar f_{{\text{cur}}}^j } \tau }} \right.
 \kern-\nulldelimiterspace} \tau }} \right)} }},i \ne j
\label{3}
\end{equation}

Correspondingly, the probability that $\bar f_{{\text{cur}}}^j ,i \ne j$ is not classified into label $i$ is $1\, - p\left( {i\left| {\bar f_{{\text{cur}}}^j } \right.} \right)$.

Assuming different embedding features being recognized as label $i$ are independent, the joint distribution of $\bar f_{{\text{cur}}}^i $
 being categorized into label $i$ and $\bar f_{{\text{cur}}}^j ,i \ne j$
 not being viewed as label $i$ is given by

\begin{equation}
P_i  = p\left( {i\left| {\bar f_{{\text{cur}}}^i } \right.} \right)\prod\limits_{j \ne i} {\left( {1\, - \,p\left( {i\left| {\bar f_{{\text{cur}}}^j } \right.} \right)} \right)} 
\label{4}
\end{equation}

The negative log likelihood is defined by,

\begin{equation}
J_i  =  - \log p\left( {i\left| {\bar f_{{\text{cur}}}^i } \right.} \right) - \sum\limits_{j \ne i} {\log \left( {1\, - \,p\left( {i\left| {\bar f_{{\text{cur}}}^j } \right.} \right)} \right)} 
\label{5}
\end{equation}

We minimize the sum of the above Eq.(4) over all the embedding features for current episodic memory, which is given by,

\begin{equation}
J_i  =  - \sum\limits_i {\log p\left( {i\left| {\bar f_{{\text{cur}}}^i } \right.} \right)}  - \sum\limits_i {\sum\limits_{j \ne i} {\log \left( {1\, - \,p\left( {i\left| {\bar f_{{\text{cur}}}^j } \right.} \right)} \right)} } 
\label{6}
\end{equation}

Then, the final loss function is formulated as follows,

\begin{equation}
L_{{\text{total}}} {\text{ = }}L_{CE} \left( {D^t ,M{\kern 1pt} ^t } \right) + \lambda J_i \left( {{\mathbf{\bar f}}_{\,{\text{pre}}}^{M_t } ,{\mathbf{\bar f}}_{\,{\text{cur}}}^{M_t } } \right)
\label{7}
\end{equation}

where $L_{CE}$ is the cross-entropy loss and $J_i$ is the regular term based on the knowledge invariant and spread-out properties. We use a parameter $\lambda $ to balance $L_{CE}$ and $J_i$. 

\subsection{Rationale Analysis}

Here, we also make a detailed rationale analysis about why minimizing Eq.(5) can realize the knowledge invariant and spread-out properties. Minimizing the above Eq.(5) can be recognized as maximizing Eq.(2) and minimizing Eq.(3) respectively. Noting that for Eq.(3), it can be rewritten as,

\begin{equation}
p\left( {i\left| {\bar f_{{\text{cur}}}^i } \right.} \right) = \frac{{\exp \left( {{{\left( {\bar f_{{\text{pre}}}^i } \right)^T \bar f_{{\text{cur}}}^i } \mathord{\left/
 {\vphantom {{\left( {\bar f_{{\text{pre}}}^i } \right)^T \bar f_{{\text{cur}}}^i } \tau }} \right.
 \kern-\nulldelimiterspace} \tau }} \right)}}
{{\exp \left( {{{\left( {\bar f_{{\text{pre}}}^i } \right)^T \bar f_{{\text{cur}}}^i } \mathord{\left/
 {\vphantom {{\left( {\bar f_{{\text{pre}}}^i } \right)^T \bar f_{{\text{cur}}}^i } \tau }} \right.
 \kern-\nulldelimiterspace} \tau }} \right) + \sum\nolimits_{k \ne i}^{} {\exp \left( {{{\left( {\bar f_{{\text{pre}}}^k } \right)^T \bar f_{{\text{cur}}}^i } \mathord{\left/
 {\vphantom {{\left( {\bar f_{{\text{pre}}}^k } \right)^T \bar f_{{\text{cur}}}^i } \tau }} \right.
 \kern-\nulldelimiterspace} \tau }} \right)} }}
\label{8}
\end{equation}

maximizing Eq.(2) means maximizing $\exp \left( {{{\left( {\bar f_{{\text{pre}}}^i } \right)^T \bar f_{{\text{cur}}}^i } \mathord{\left/
 {\vphantom {{\left( {\bar f_{{\text{pre}}}^i } \right)^T \bar f_{{\text{cur}}}^i } \tau }} \right.
 \kern-\nulldelimiterspace} \tau }} \right)$ and minimizing $\exp \left( {{{\left( {\bar f_{{\text{pre}}}^k } \right)^T \bar f_{{\text{cur}}}^i } \mathord{\left/
 {\vphantom {{\left( {\bar f_{{\text{pre}}}^k } \right)^T \bar f_{{\text{cur}}}^i } \tau }} \right.
 \kern-\nulldelimiterspace} \tau }} \right),k \ne i$. Maximizing $\exp \left( {{{\left( {\bar f_{{\text{pre}}}^i } \right)^T \bar f_{{\text{cur}}}^i } \mathord{\left/
 {\vphantom {{\left( {\bar f_{{\text{pre}}}^i } \right)^T \bar f_{{\text{cur}}}^i } \tau }} \right.
 \kern-\nulldelimiterspace} \tau }} \right)$ can increase the cosine similarity between $\bar f_{{\text{pre}}}^i $ and $\bar f_{{\text{cur}}}^i $ where $\bar f_{{\text{pre}}}^i $ and $\bar f_{{\text{cur}}}^i $ are normalized features extracted by the previous model and those by the current one, respectively, which can encourage the knowledge learned by the current model to be consistent to those by the previous model. On the other hand, minimizing $\exp \left( {{{\left( {\bar f_{{\text{pre}}}^k } \right)^T \bar f_{{\text{cur}}}^i } \mathord{\left/
 {\vphantom {{\left( {\bar f_{{\text{pre}}}^k } \right)^T \bar f_{{\text{cur}}}^i } \tau }} \right.
 \kern-\nulldelimiterspace} \tau }} \right),k \ne i$ requires $\bar f_{{\text{cur}}}^i $ and other features of the samples are separated. Considering all the instances within the current episodic memory, the embedding features are compelled to be separated from each other, which can lead to the spread-out property. 

In a similar manner, Eq.(3) can be rewritten as, 

\begin{equation}
p\left( {i\left| {\bar f_{{\text{cur}}}^j } \right.} \right) = \frac{{\exp \left( {{{\left( {\bar f_{{\text{pre}}}^i } \right)^T \bar f_{{\text{cur}}}^j } \mathord{\left/
 {\vphantom {{\left( {\bar f_{{\text{pre}}}^i } \right)^T \bar f_{{\text{cur}}}^j } \tau }} \right.
 \kern-\nulldelimiterspace} \tau }} \right)}}
{{\exp \left( {{{\left( {\bar f_{{\text{pre}}}^j } \right)^T \bar f_{{\text{cur}}}^j } \mathord{\left/
 {\vphantom {{\left( {\bar f_{{\text{pre}}}^j } \right)^T \bar f_{{\text{cur}}}^j } \tau }} \right.
 \kern-\nulldelimiterspace} \tau }} \right) + \sum\nolimits_{k \ne j}^{} {\exp \left( {{{\left( {\bar f_{{\text{pre}}}^k } \right)^T \bar f_{{\text{cur}}}^j } \mathord{\left/
 {\vphantom {{\left( {\bar f_{{\text{pre}}}^k } \right)^T \bar f_{{\text{cur}}}^j } \tau }} \right.
 \kern-\nulldelimiterspace} \tau }} \right)} }},i \ne j
\label{9}
\end{equation}

Note that the inner product $\left( {\bar f_{{\text{pre}}}^j } \right)^T \bar f_{{\text{cur}}}^j $ tends to 1 and the scale factor $\tau $ generally is set to a small value (set to 0.1 in our experiment). Thus, the term $\exp \left( {{{\left( {\bar f_{{\text{pre}}}^j } \right)^T \bar f_{{\text{cur}}}^j } \mathord{\left/
 {\vphantom {{\left( {\bar f_{{\text{pre}}}^j } \right)^T \bar f_{{\text{cur}}}^j } \tau }} \right.
 \kern-\nulldelimiterspace} \tau }} \right)$ generally dominates the value of the whole denominator. Minimizing Eq.(3) encourages the value of$\exp \left( {{{\left( {\bar f_{{\text{pre}}}^i } \right)^T \bar f_{{\text{cur}}}^j } \mathord{\left/
 {\vphantom {{\left( {\bar f_{{\text{pre}}}^i } \right)^T \bar f_{{\text{cur}}}^j } \tau }} \right.
 \kern-\nulldelimiterspace} \tau }} \right)$ should be minimized, which aims at increasing the dissimilarity between $\bar f_{{\text{pre}}}^i $ and $\bar f_{{\text{cur}}}^j $. Therefore, it would further encourage the spread-out property and then be beneficial to transfer the structural knowledge underlying in embedding features for previous tasks to current task, such as the surrounding of instance embedding features, inter-instance location relationships and inter-instance geometric structure information.

\section{Experimental evaluations}

In this section, we compare our OCLKISP with other baselines on several datasets used in continual learning to verify the effectiveness of our proposed method. Then, we introduce the metrics and implementation details used in our experiments. Finally, we also perform ablation studies to validate the improvement of my proposed method.

\subsection{Experiment Setting}

\textbf{Datasets}	Following the protocol defined in 
\cite{18lopez2017gradient}, we evaluate the performance of my proposed method and baselines on Split SVHN, Split CIFAR100, Split CUB200, Split TinyImageNet200.

\begin{itemize}
\item[1)]\textbf{Split SVHN
\cite{53rajasegaran2020itaml}.} The SVHN
\cite{54netzer2011reading} consists of 600,000 images with the size of $32 \times 32$ pixels with 10 classes. In our experiments, we part this dataset into 5 disjoint tasks and every task has no overlap classes.

\item[2)]\textbf{Split CIFAR100
\cite{18lopez2017gradient}.} CIFAR100
\cite{55krizhevsky2009learning} contains 60000 images with the size of $32 \times 32$ pixel in 100 object categories, where 50000 images are applied to train a model and others are used to test. In this work, we divide this dataset into 20 disjoint tasks and each task contains 5 object categories.

\item[3)]\textbf{Split CUB200
\cite{56wah2011caltech}.} This benchmark is a fine-grained image dataset with 200 bird species. This dataset includes 11788 images in total. Among 11788, 5994 images are the training data and 5794 images are the testing data. Here, we divide CUB200 into 20 disjoint subsets and every subset is recognized as a task.

\item[4)]\textbf{Split Tiny-Image-Net200
\cite{57gupta2020maml}.} This dataset
\cite{58le2015tiny} contains 120,000 images with the size of   pixels in 200 classes and we divide 200 object categories into 40 5-way classification tasks in this paper.
\end{itemize} 

\textbf{Baselines}	 We extensively compare our OCLKISP with following several state-of-the-art continual learning approaches, including \textbf{Finetune:} we learn this model without any regularization and episodic memory; \textbf{LwF}(Learning without Forgetting)
\cite{14li2016learning}: this method apply knowledge distillation to resist catastrophic forgetting; \textbf{EWC}(Elastic Weight Consolidation)
\cite{16kirkpatrick2017overcoming}: this method constrain the change of the important parameters via Fisher Information Matrix to learn continually a sequence of tasks; \textbf{ICARL}(Incremental Classifier and Representation Learning)
\cite{39rebuffi2017icarl}: this model apply nearest center mean as classifier at inference and knowledge distillation is used to avoid forgetting; \textbf{GEM}(Gradient Episodic Memory)
\cite{18lopez2017gradient}:an episodic memory is used to avoid forgetting and positive backward transfer is considered in this model; \textbf{AGEM}(Average Gradient Episodic Memory)
\cite{40chaudhry2019efficient}: this model apply the average gradient information of examples in the episodic memory to control the change of the model; \textbf{ER}(Experience Replay)
\cite{59chaudhry2019continual}: In this method, experience replay is used to alleviate forgetting; \textbf{DER and DER++} (Dark Experience Replay)
\cite{60buzzega2020dark}: These two approaches address forgetting through mixing rehearsal with knowledge distillation, thus promoting consistency with its past while alleviating the negative effects due to data imbalance. Note that DER++ equips the objective function of DER with an additional term on buffer datapoints.

\textbf{Metrics}	 For comprehensive evaluations, we adopt three standard metrics: The Final Accuracy (FA) 
\cite{18lopez2017gradient}, The Global Average Accuracy (GA)
\cite{61diaz2018don}, Forgetting Measure (FM) 
\cite{40chaudhry2019efficient}, and Learning Accuracy (LA)
\cite{21riemer2018learning}. FA denotes the final average classification accuracy of all observed tasks $\left\{ {1,2, \cdots ,t, \cdots ,T} \right\}$ after training the whole sequence; Different from FA, GA is an accuracy metric that takes into account the performance of a model at each task phase $t$ better characterizes the dynamic aspects of continual learning. GA and FA are used to measure the performance of the final model from different perspectives. FM indicates the influence that learning a new task has the performance on prior tasks, in other words, the average forgetting of all previous tasks after learning a new task; LA reflects the performance of a model on a task right after it finishes learning that task, which can indicate the ability to learn new knowledge to some extent.

Here, given the train-test accuracy matrix $a_{i,j} $, which denotes the test classification accuracy of the model on task  $t_j $ after observing the last instance about the current task $t_i $;$T$ is the total number of tasks in the current task sequence. Thus, the FA, GA, FM, and LA are defined as follows:

\begin{equation}
{\text{The Final Accuracy (FA) :    ACC}} = \frac{1}
{T}\sum\limits_{i = 1}^T {a_{T,i} } {\text{      }}
\label{10}
\end{equation}

\begin{equation}
{\text{The Global Accuracy (GA) :    GA}} = \frac{{\sum\limits_{i \geqslant j}^T {a_{i,j} } {\text{ }}}}
{{{{T(T + 1)} \mathord{\left/
 {\vphantom {{T(T + 1)} 2}} \right.
 \kern-\nulldelimiterspace} 2}}}{\text{     }}
\label{11}
\end{equation}

\begin{equation}
{\text{Forgetting Measure (FM) :    FM}} = \frac{1}
{{T - 1}}\sum\limits_{j = 1}^{T - 1} {\mathop {\max }\limits_{l < T} a_{l,j}  - a_{T,j} } 
\label{12}
\end{equation}

\begin{equation}
{\text{Learning Accuracy (LA) :    LA}} = \frac{1}
{T}\sum\limits_{i = 1}^T {a_{i,i} } {\text{  }}
\label{12}
\end{equation}

\textbf{Implementation Details}	In this work, we follow the implementation details proposed in the literature
\cite{40chaudhry2019efficient}. In particular, during the training process, we also use the cross-validation principle via bilevel optimization
\cite{62colson2007overview} to cross-validate the parameter of our model and perform online continual learning 
\cite{52pham2020bilevel,63nichol2018first}. In addition, because we insist that the learning process is a “single pass through data” which means that the model is only trained for one epoch for each task in all of our experiments. For each method, we use the same hyper-parameter configuration and protocol as provided in the corresponding original paper. We set each data batch size to 10 for all methods. Here, for comparisons of memory-based methods, such as GEM, A-GEM, ER, follow 
\cite{40chaudhry2019efficient,52pham2020bilevel}, the size of episodic memory for every task is set to 65, 25, 65 and 128 on Split CIFAR10, Split SVHN, Split CIFAR100 and Split Tiny-ImageNet respectively.
 
Specifically, for SVHN, CIFAR, Tiny-Image-Net, we use a reduced ResNet18 as backbone, and details of network architecture are shown in the following Fig.4. For CUB200, we use a pretrained full ResNet18
\cite{64he2016deep} as the baseline architecture. We apply SGD to train our model with the minibatch size of 10 in an online learning manner. For SVHN and CUB200 datasets, one gradient update is sufficient. Conversely, for CIFAR100 and Tiny-ImageNet, we perform a few iterations over a batch and the number of iterations is set to 2 and 3 respectively, as done by works 
\cite{mai2022online,NEURIPS2019_15825aee}
 Moreover, we select the ring buffer proposed in 
\cite{18lopez2017gradient} as our memory writing strategy. Lastly, for each experimental result, the experiments are conducted five times, and the average FA, GA, FM, and LA are reported in this work.

\begin{figure}[H]
	\centering
	\includegraphics[width=\textwidth]{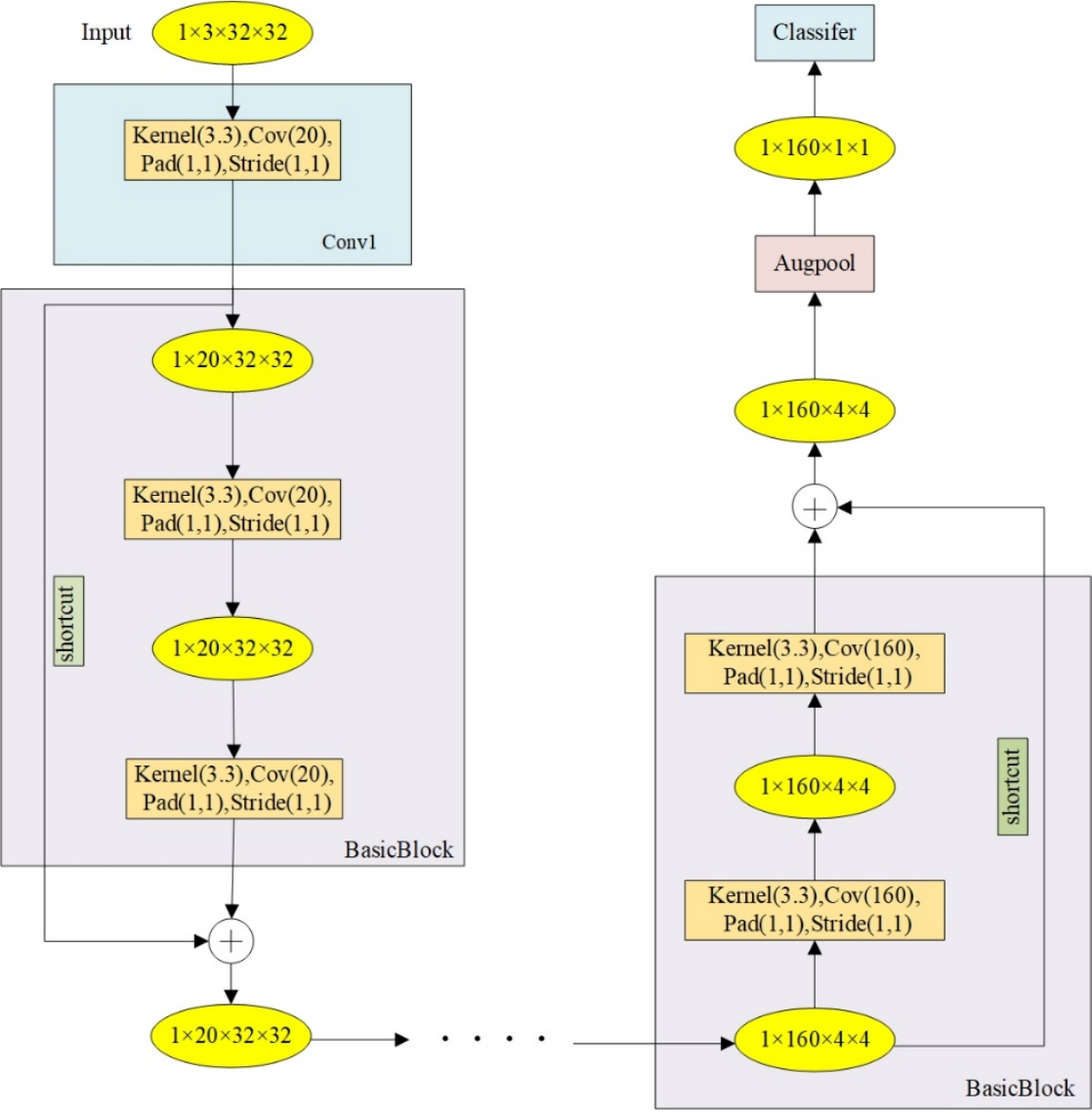}
	\caption{Details of model architectures used in this work on CIFAR dataset}
	\label{fig5}
\end{figure}

\subsection{Results and Analysis}

In this subsection, we evaluate our OCLKISP method’s performance by comparing it with other state-of-the-art models in terms of FA, GA, FM, and LA with quantitative and qualitative results. Furthermore, in order to verify the effectiveness of our proposed method, we also conduct ablation studies. The specific experimental results and analysis are shown below.

\subsubsection{OCLKISP Performance on different datasets}

\textbf{1) Different Evaluation Metrics on various datasets}

The numerical results of different baselines are reported in Table 2 on four popular benchmarks including Split Cifar100(20T), Split SVHN(5T), Split CUB200(20T), and Split Tiny-Image-Net (40T). Across all approaches, the higher FA and GA indicate the better performance of the model; The value of LA reflects the ability to learn new knowledge, higher is better; Then, FM is used to measure the forgetting of previous knowledge, and this value is higher, catastrophic forgetting is more prominent. 

All numerical results are reported in table 2. From the experimental results, we make several observations. Firstly, we can see that our OCLKISP method outperforms all the baselines by a large margin across all the benchmarks consistently, for example, our proposed method achieves the best FA, LA and GA on all dataset. Besides, consistent with the finding of previous studies, the regularization methods, such as EWC, LWF, exhibit relatively poor performance and catastrophic forgetting of the previously learned knowledge. Compared with regularization-based methods, the ICARL algorithm makes a further improvement via the nearest-example algorithm. Then, we also note that all of the considered memory-based approaches, such as ER, GEM, AGEM, were able to obtain competitive performance on all datasets. 

\begin{table}[H]
\centering
\caption{The results of different baselines on different datasets, i.e., 95\% confidence intervals reported after the symbol ±.}
\begin{tabular}{cccccc}
\hline
Methods  & Dataset                          & FA                  & GA                  & LA                  & FM                  \\ 
\hline
Finetune & \multirow{10}{*}{Split Cifar100} & 34.93±0.99          & 41.90±2.02          & 66.01±0.32          & 32.74±0.96          \\
EWC      &                                  & 38.24±3.62          & 42.75±3.20          & 64.38±0.73          & 26.86±3.74          \\
LWF      &                                  & 42.24±1.19          & 44.61±1.60          & 64.25±0.61          & 22.40±1.00          \\
ICARL    &                                  & 46.48±0.47          & 49.69±1.03          & 66.01±1.56          & 18.77±1.66          \\
GEM      &                                  & 62.63±0.58          & 61.18±1.04          & 67.99±0.55          & 7.08±0.41           \\
AGEM     &                                  & 55.59±1.43          & 57.08±0.61          & 67.56±0.84          & 13.27±0.81          \\
ER       &                                  & 63.04±1.56          & 62.29±0.76          & 68.73±0.84          & 5.83±2.17           \\
DER      &                                  & 65.10±0.65          & 63.92±0.67          & 69.53±0.99          & 6.11±0.31           \\
DER++    &                                  & 67.84±1.14          & 64.59±3.07          & 69.33±0.49          & 3.47±0.46           \\
OCLKISP  &                                  & \textbf{68.70±1.00} & \textbf{64.81±2.23} & \textbf{70.47±0.44} & \textbf{2.81±0.42}  \\
\hline
\end{tabular}
\end{table}

\begin{table}[H]
\centering
\begin{tabular}{cccccc} 
\hline
Methods  & Dataset                                & FA                  & GA                  & LA                  & FM                  \\ 
\hline
Finetune & \multirow{10}{*}{Split CUB200}         & 34.93±1.27          & 41.90±1.06          & 72.30±1.60          & 12.69±2.23          \\
EWC      &                                        & 65.45±1.46          & 67.87±1.84          & 74.52±0.71          & 11.82±1.48          \\
LWF      &                                        & 66.43±2.31          & 67.59±2.46          & 75.64±1.53          & 11.19±2.50          \\
ICARL    &                                        & 67.59±1.55          & 73.45±3.90          & 76.55±0.39          & 9.81±1.62           \\
GEM      &                                        & 81.69±1.57          & 79.39±1.68          & 77.01±1.81          & 2.23±1.57           \\
AGEM     &                                        & 73.77±0.25          & 73.45±1.12          & 75.96±0.57          & 5.16±0.37           \\
ER       &                                        & 81.42±1.07          & 81.05±0.51          & 78.67±0.74          & 3.59±0.93           \\
DER      &                                        & 83.17±0.43          & 80.46±0.82          & 76.59±1.30          & \textbf{1.15±0.79}  \\
DER++    &                                        & 83.32±0.46          & 81.02±0.51          & 77.56±1.41          & 1.22±0.56           \\
OCLKISP  &                                        & \textbf{85.08±0.51} & \textbf{83.45±0.73} & \textbf{83.22±0.69} & 1.62±0.49           \\ 
\hline
Finetune & \multirow{10}{*}{Split Tiny-Image-Net} & 34.10±2.80          & 35.52±3.98          & 64.82±0.49          & 31.54±3.01          \\
EWC      &                                        & 35.09±1.57          & 35.49±4.15          & 65.13±1.16          & 30.88±1.11          \\
LWF      &                                        & 36.41±0.30          & 35.73±4.38          & 64.77±0.23          & 29.12±0.36          \\
ICARL    &                                        & 43.87±0.31          & 45.89±3.08          & 66.19±0.54          & 23.04±0.57          \\
GEM      &                                        & 63.24±0.78          & 61.76±2.27          & 67.00±0.63          & 6.46±0.43           \\
AGEM     &                                        & 60.37±0.79          & 57.17±1.98          & 66.75±0.38          & 7.92±0.84           \\
ER       &                                        & 67.44±0.19          & 63.36±2.86          & 68.48±0.33          & 3.65±0.36           \\
DER      &                                        & 67.50±0.43          & 64.05±3.08          & 69.29±0.78          & 4.58±0.17           \\
DER++    &                                        & 69.43±0.37          & 66.78±3.04          & 71.31±0.41          & 4.79±0.19           \\
OCLKISP  &                                        & \textbf{69.96±0.54} & \textbf{67.55±2.19} & \textbf{71.91±0.35} & \textbf{3.49±0.39}  \\ 
\hline
Finetune & \multirow{10}{*}{Split SVHN}           & 75.87±7.18          & 79.34±8.15          & 96.07±0.10          & 25.25±7.06          \\
EWC      &                                        & 82.65±1.52          & 81.42±6.86          & 94.39±1.36          & 14.66±3.48          \\
LWF      &                                        & 83.27±1.53          & 79.50±7.06          & 93.97±0.87          & 13.37±2.86          \\
ICARL    &                                        & 86.89±1.59          & 85.41±4.74          & 95.39±0.81          & 10.62±2.77          \\
GEM      &                                        & 90.00±1.25          & 89.76±2.68          & 95.72±0.79          & 7.14±2.41           \\
AGEM     &                                        & 90.84±1.43          & 89.81±2.15          & 95.26±0.52          & 5.54±1.67           \\
ER       &                                        & 91.57±1.07          & 91.72±2.05          & 96.43±0.32          & 6.08±1.43           \\
DER      &                                        & 93.02±0.71          & 92.42±1.81          & 95.88±0.47          & 3.67±1.17           \\
DER++    &                                        & 93.44±1.16          & 93.29±1.31          & 96.26±0.15          & 3.40±1.84           \\
OCLKISP  &                                        & \textbf{93.55±0.28} & \textbf{93.66±1.51} & \textbf{96.60±0.25} & \textbf{3.81±0.32}  \\
\hline
\end{tabular}
\end{table}

\textbf{2) The evolution of the final average accuracy as the number of tasks increases}

As new tasks are learned, the evolution of the final accuracies and global accuracies in different datasets are shown in Figs. 6 and 7 respectively. Each curve illustrates the different methods’ final accuracies or global accuracies for all the past tasks in the current training phase. Firstly, we can find that the OCLKISP method proposed in this work outperforms all the baselines by a significant margin across the various benchmarks consistently, which would further verify the above observations. Then, we also observe that the curves of traditional methods, such as finetune, LWF, and EWC, declines dramatically as more tasks come, which means, these methods have limited ability to resist catastrophic forgetting. Moreover, the downward trend of those memory-based approaches is relatively slow, which indicates that these methods with the experience replay strategy can more effectively alleviate forgetting than regularization-based methods. It is also worth to be highlighted that as more tasks are learned, our OCLKISP method enjoys superior performance than other baselines, which illustrates that the OCLKISP method exhibits a superior ability to prevent forgetting in a continual learning setting.

\begin{figure}[H]
\centering
\subfigure[Split SVHN]{\label{fig:subfig:a}
\includegraphics[width=0.35\linewidth]{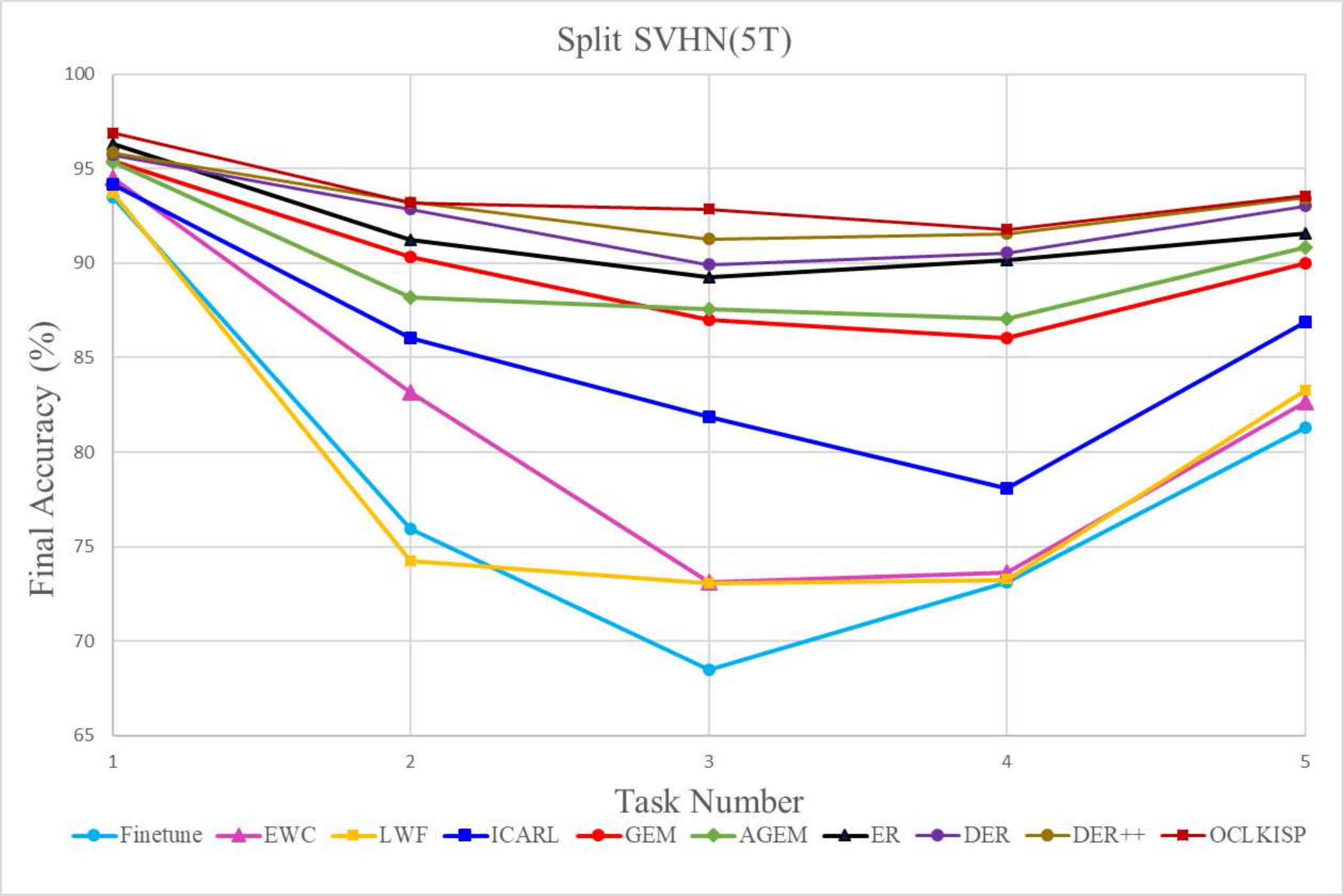}}
\hspace{0.01\linewidth}
\subfigure[Split CUB200]{\label{fig:subfig:b}
\includegraphics[width=0.35\linewidth]{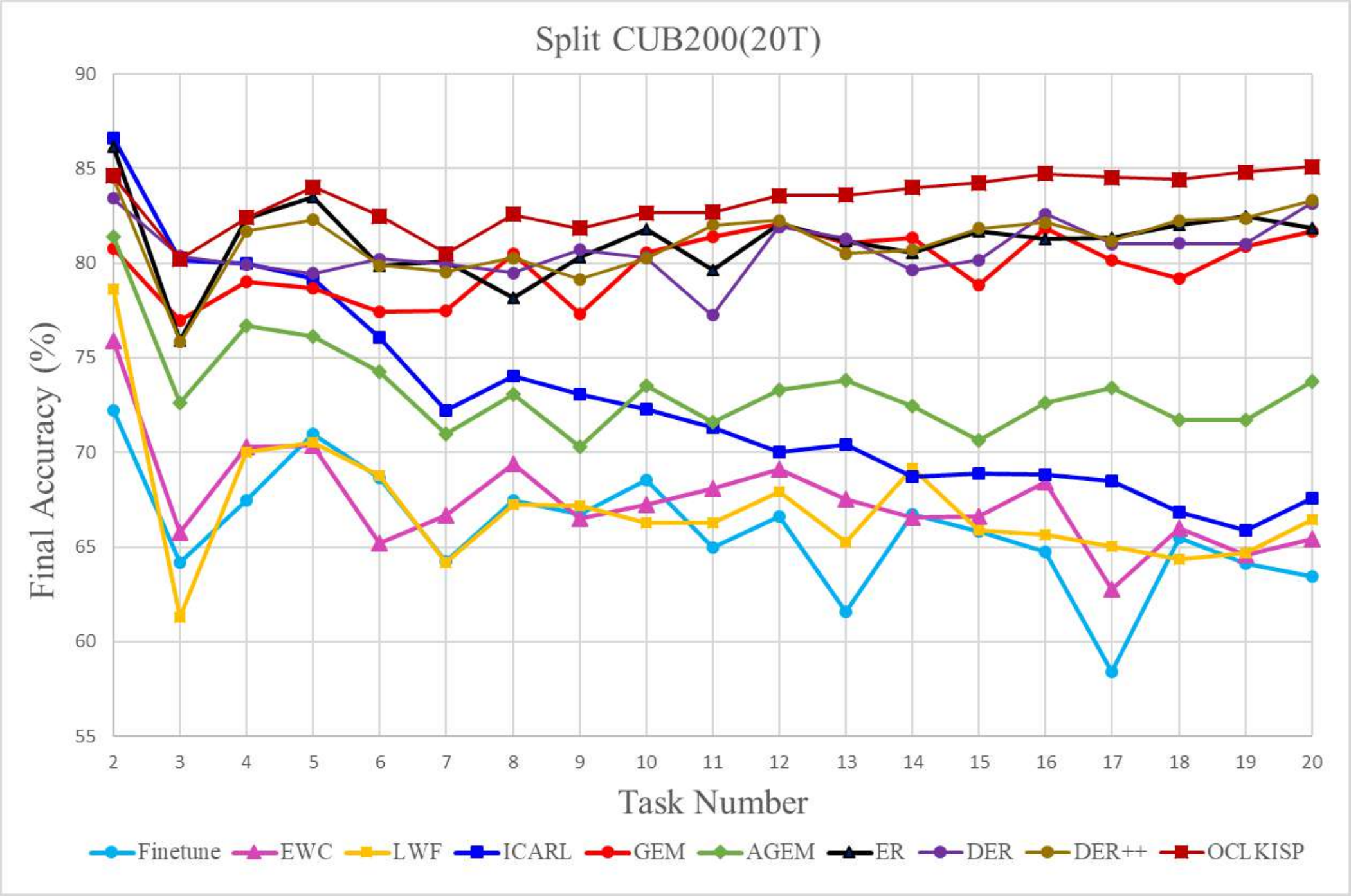}}
\vfill
\subfigure[Split Cifar100]{\label{fig:subfig:a}
\includegraphics[width=0.35\linewidth]{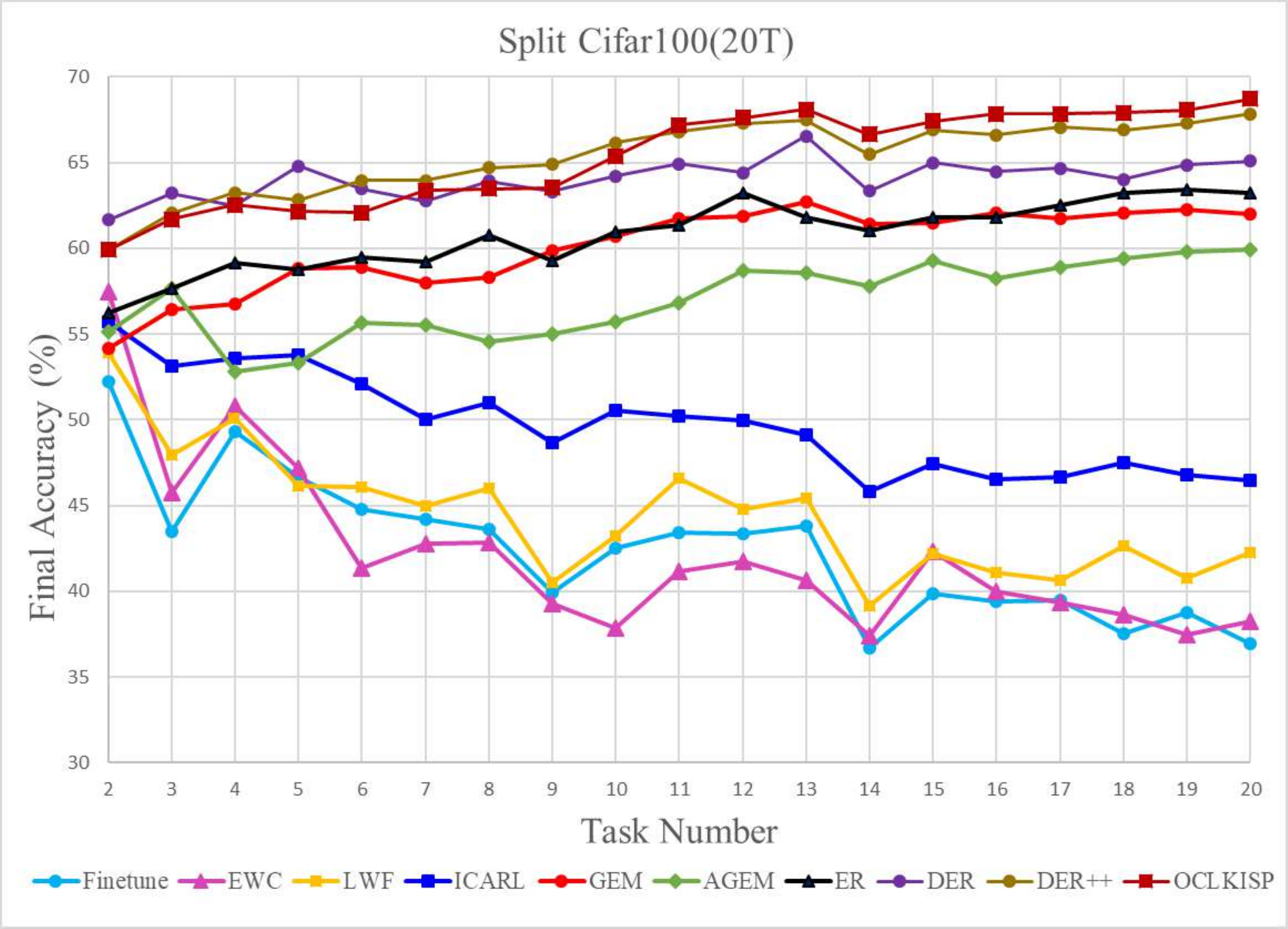}}
\hspace{0.01\linewidth}
\subfigure[Split Tiny-ImageNet]{\label{fig:subfig:b}
\includegraphics[width=0.35\linewidth]{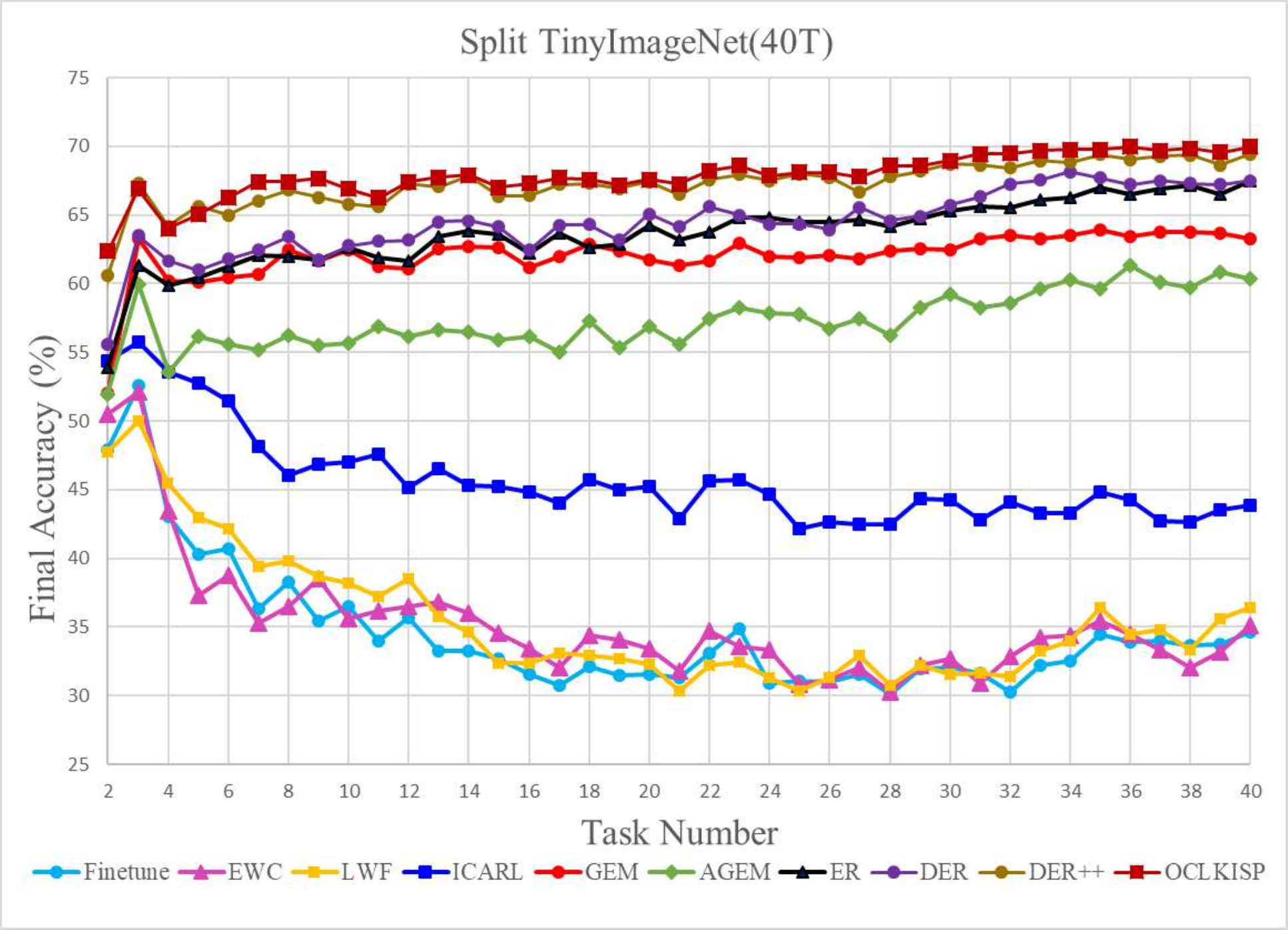}}
\caption{Evolution curves of the final accuracies as new tasks arrive on different datasets}
\label{fig6}
\end{figure}

\begin{figure}[H]
\centering
\subfigure[Split SVHN]{\label{fig:subfig:a}
\includegraphics[width=0.35\linewidth]{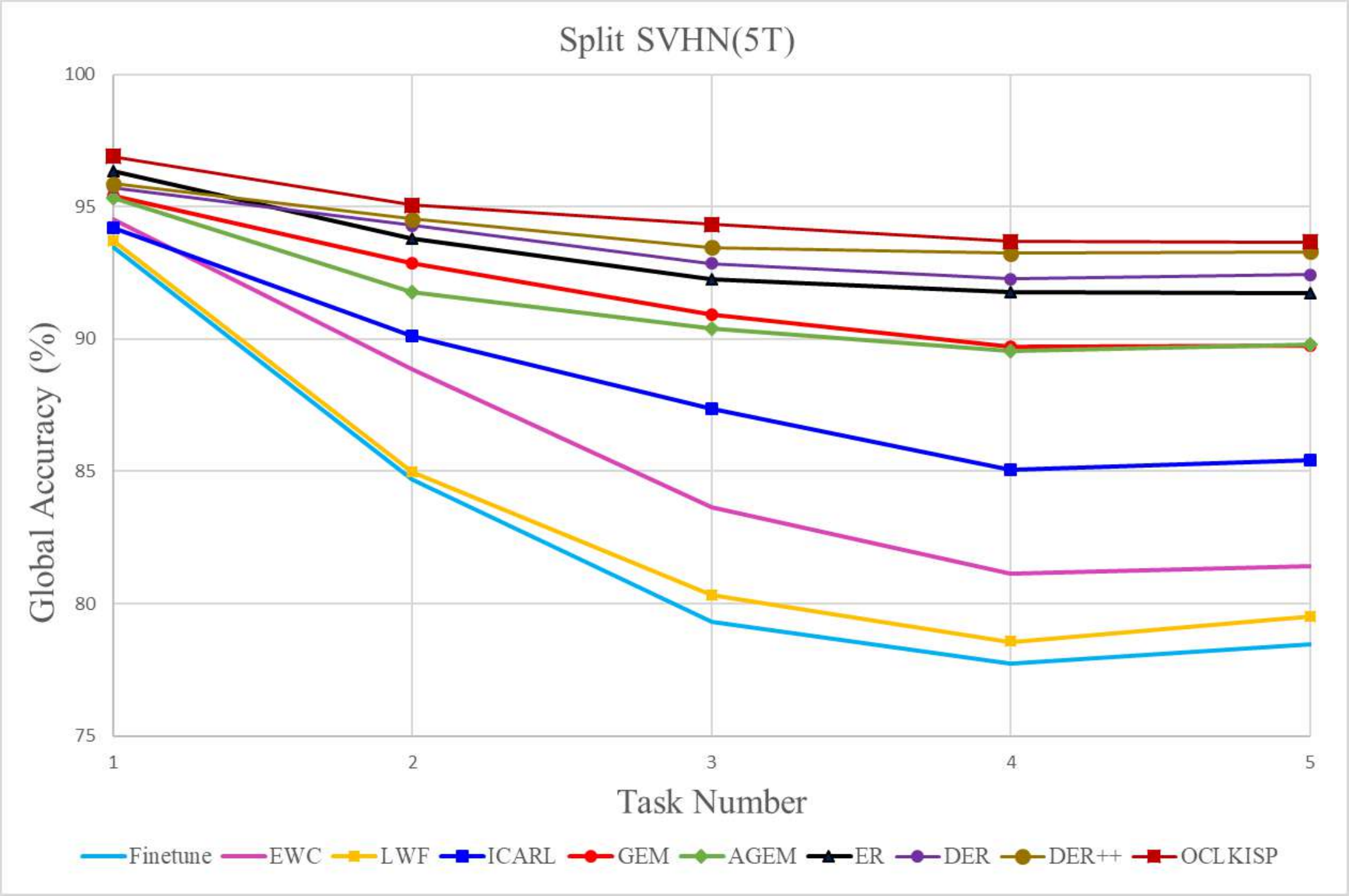}}
\hspace{0.01\linewidth}
\subfigure[Split CUB200]{\label{fig:subfig:b}
\includegraphics[width=0.35\linewidth]{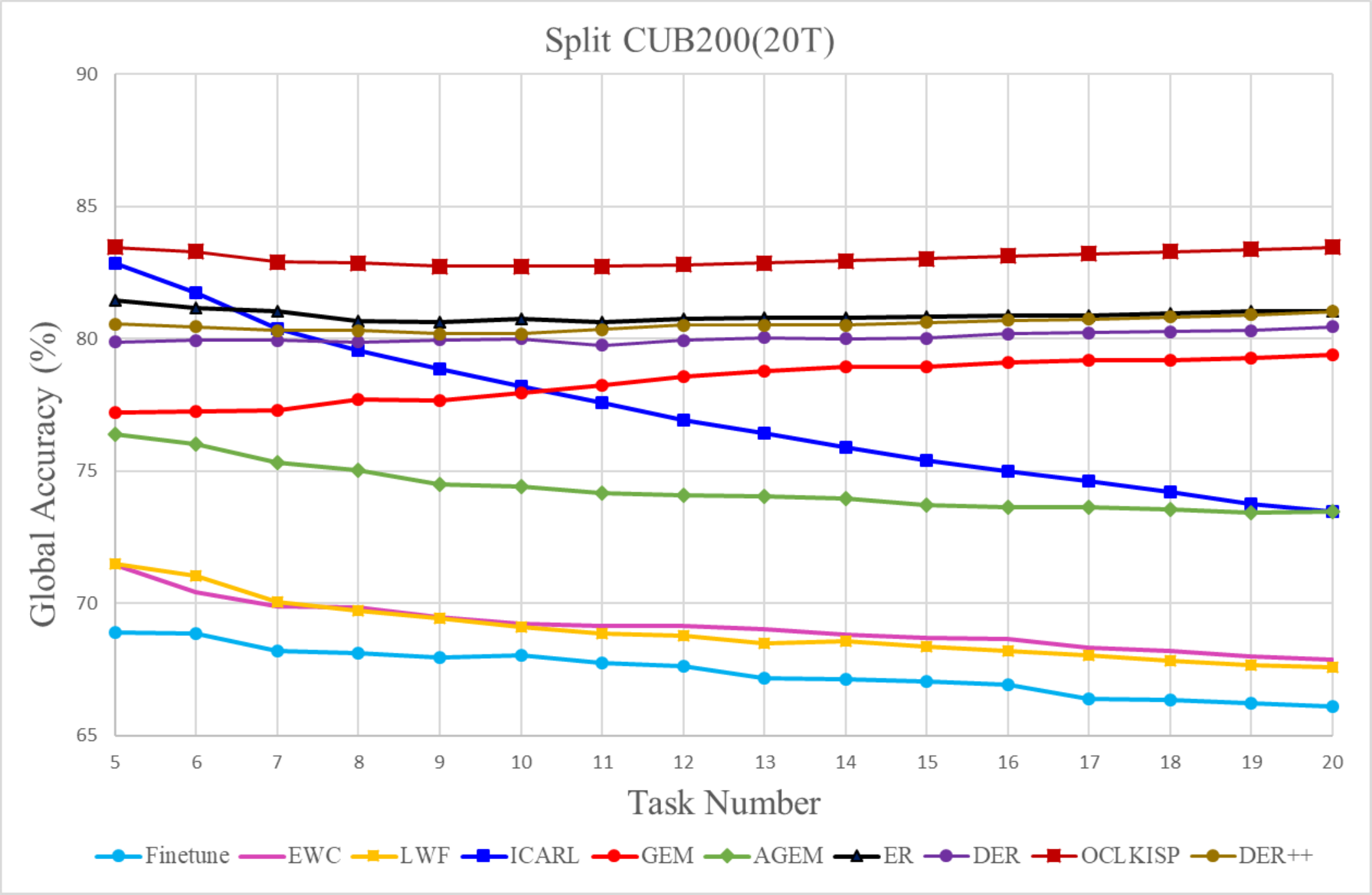}}
\vfill
\subfigure[Split Cifar100]{\label{fig:subfig:a}
\includegraphics[width=0.35\linewidth]{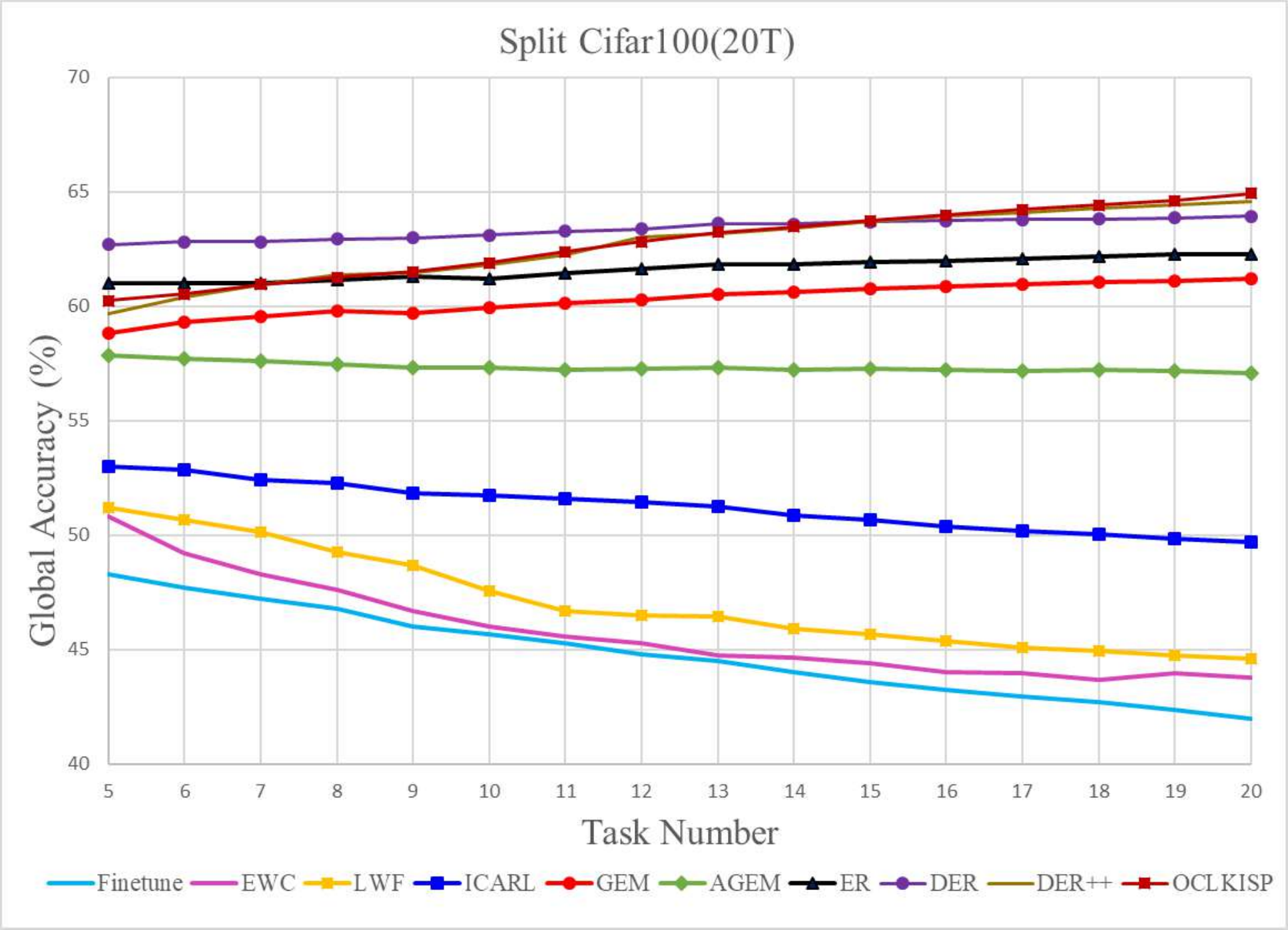}}
\hspace{0.01\linewidth}
\subfigure[Split Tiny-ImageNet]{\label{fig:subfig:b}
\includegraphics[width=0.35\linewidth]{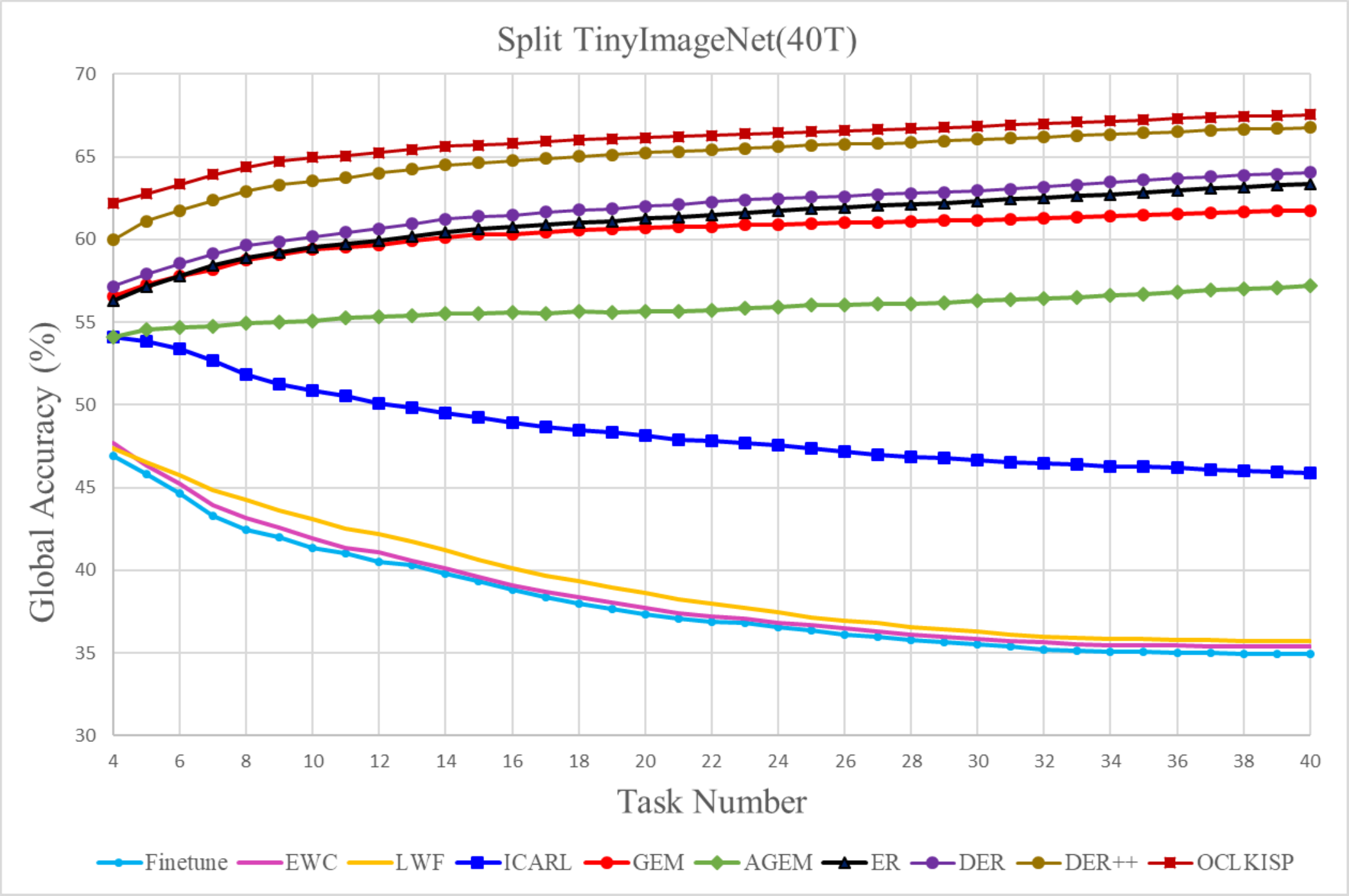}}
\caption{Evolution curves of the global accuracies as new tasks arrive on different datasets}
\label{fig7}
\end{figure}

\textbf{3)	The impact of KISP against other baselines}

In this section, we compare our OCLKISP method with two variations of baselines and further analyze the impact of the KISP. The different methods are explained as follows: \textbf{w/o KISP}: our model is trained without KISP; \textbf{OCL-LFC:} Compared to w/o KISP, the Less-Forget Constraint(LFC) 
\cite{24hou2019learning} is employed which can encourage the orientation of features extracted via current model to be consistent to those by previous model, and then we can prevent the influence due to the learning bias; \textbf{OCL-RLD}
\cite{25ni2021alleviate}: this method, called representation-level distillation(RLD) prevent the learning bias by guiding the representation of the episodic memory on current model close to the previous model. 

The experimental results are shown in Table 2. With the help of the LFC and RLD, OCL-LFC and OCL-RLD are better than w/o KISP which illustrates those can alleviate the influence of the bias to some extent, such as the final accuracy on Split CIFAR100 is boosted from 66.03\% to 67.60\% and 68.59\% respectively. However, those methods neglect the important structural knowledge of previous tasks, i.e., similarities or dissimilarities between the different embedding features. Based on this, in this work, we design a novel loss via the knowledge invariant and spread-out properties. In our OCLKISP method, the knowledge invariant is used to alleviate the forgetting due to the learning bias. Compared to LFC and RLD, our proposed method can further transfer the structure knowledge from prior tasks and then achieve slightly better accuracies than other models, such as the higher FA on all datasets. Note that OCLKISP obtains worse global accuracy than OCL-LFC on the Split CUB200 dataset. This result means the final accuracies of our OCLKISP are not always better than OCL-LFC at each task phase when the task  comes sequentially. However, our proposed model has a higher FA than OCL-LFC which means that when all tasks are learned, our OCLKISP enjoys a better final performance than other models. The following results validate the effectiveness of our proposed method.

\begin{table}[H]
\centering
\caption{The numerical results on different datasets, i.e., 95\% confidence intervals reported after the symbol ±}
\begin{tabular}{llllll} 
\hline
Methods  & Datasets                              & FA                  & GA                  & LA                  & FM                  \\ 
\hline
w/o KISP & \multirow{4}{*}{Split Cifar100}       & 66.03±1.34          & 63.30±1.70          & 69.20±0.62          & 4.56±0.94           \\
OCL-LFC  &                                       & 67.60±0.49          & 64.07±2.85          & 69.36±0.52          & \textbf{2.33±1.68}  \\
OCL-RLD  &                                       & 68.59±0.19          & 64.59±2.36          & 69.77±0.64          & 2.99±0.45           \\
OCLKISP  &                                       & \textbf{68.70±1.00} & \textbf{64.81±2.23} & \textbf{70.47±0.44} & 2.81±0.42           \\ 
\hline
w/o KISP & \multirow{4}{*}{Split CUB200}         & 84.31±0.48          & 83.14±0.24          & 82.58±0.76          & 1.91±0.59           \\
OCL-LFC  &                                       & 84.48±0.24          & \textbf{84.00±0.45} & 83.21±0.54          & 2.45±0.49           \\
OCL-RLD  &                                       & 84.45±0.52          & 83.05±0.57          & 82.33±0.36          & 1.84±0.43           \\
OCLKISP  &                                       & \textbf{85.08±0.51} & 83.45±0.73          & \textbf{83.22±0.69} & \textbf{1.62±0.49}  \\ 
\hline
w/o KISP & \multirow{4}{*}{Split SVHN}           & 92.44±1.32          & 93.11±1.55          & 96.42±0.22          & 5.06±1.77           \\
OCL-LFC  &                                       & 92.88±0.74          & 93.35±1.65          & 96.58±0.10          & 4.62±0.91           \\
OCL-RLD  &                                       & 92.24±0.68          & 92.60±2.12          & \textbf{96.63±0.19} & 5.51±0.87           \\
OCLKISP  &                                       & \textbf{93.55±0.28} & \textbf{93.66±1.51} & 96.60±0.25          & \textbf{3.81±0.32}  \\ 
\hline
w/o KISP & \multirow{4}{*}{Split Tiny-Image-Net} & 68.89±0.53          & 66.99±1.75          & 71.75±0.61          & 4.28±0.62           \\
OCL-LFC  &                                       & 69.01±0.30          & 66.77±2.32          & 71.49±0.20          & 3.94±0.11           \\
OCL-RLD  &                                       & 69.05±0.49          & 66.89±1.74          & \textbf{71.96±0.88} & 3.90±0.22           \\
OCLKISP  &                                       & \textbf{69.96±0.54} & \textbf{67.55±2.19} & 71.91±0.35          & \textbf{3.49±0.39}  \\
\hline
\end{tabular}
\end{table}

\subsubsection{Ablation study}

In this work, we propose a novel OCLKISP method and during the training process in continual learning, OCLKISP can capture the beneficial structure knowledge of the embedding features in episodic memory via the knowledge invariant and spread-out properties and then enhance the transfer of knowledge from old tasks. To provide more in-depth insight into the working mechanism of our proposed method, we conduct additional ablative experiments in which we further discuss the importance of knowledge invariant and spread-out properties (KISP) for our proposed method.
 
Table 3 and Fig. 8 show the experimental results of the ablation studies, where w/o KISP denotes the method removing the constraint of knowledge invariant and spread-out properties on our OCLKISP method. By comparing the results of Table 3, we observe that our OCLKISP method without KISP exhibits a worse performance than OCLKISP on all datasets, for example, there is a 2.67\% decline on the Split CIFAR100 dataset in terms of FA. Thus, we argue that the idea of KISP is indeed beneficial to enhancing the transfer of knowledge learned in past tasks and then alleviating forgetting in the continual learning environment. Furthermore, in order to further analyze the performance variation from a global accuracy view, Fig. 8 shows the performance variation of GA as new tasks arrive. We can observe that on SVHN and CUB200 datasets, our OCLKISP method achieves superior performance than the w/o KISP method consistently. Furthermore, it is noted that in all cases, compared with the w/o KISP method, our OCLKISP method offers a more significant improvement when the number of tasks increases. These results validate the effectiveness of our proposed method, which also shows that our proposed OCLKISP is a promised approach in the context of online continual learning.

\begin{table}[H]
\centering
\caption{The numerical results of different ablation studies, i.e., 95\% confidence intervals reported after the symbol ±.}
\begin{tabular}{ccccccc} 
\hline
Methods  & Datasets                              & FA                  & GA                  & LA                  & FM                  \\ 
\hline
w/o KISP & \multirow{2}{*}{Split Cifar100}       & 66.03±1.34          & 63.30±1.70          & 69.20±0.62          & 4.56±0.94           \\
OCLKISP  &                                       & \textbf{68.70±1.00} & \textbf{64.81±2.23} & \textbf{70.47±0.44} & \textbf{2.81±0.42}  \\ 
\hline
w/o KISP & \multirow{2}{*}{Split CUB200}         & 84.31±0.48          & 83.14±0.24          & 82.58±0.76          & 1.91±0.59           \\
OCLKISP  &                                       & \textbf{85.08±0.51} & \textbf{83.45±0.73} & \textbf{83.22±0.69} & \textbf{1.62±0.49}  \\ 
\hline
w/o KISP & \multirow{2}{*}{Split SVHN}           & 92.44±1.32          & 93.11±1.55          & 96.42±0.22          & 5.06±1.77           \\
OCLKISP  &                                       & \textbf{93.55±0.28} & \textbf{93.66±1.51} & \textbf{96.60±0.25} & \textbf{3.81±0.32}  \\ 
\hline
w/o KISP & \multirow{2}{*}{Split Tiny-Image-Net} & 68.89±0.53          & 66.99±1.75          & 71.75±0.61          & 4.28±0.62           \\
OCLKISP  &                                       & \textbf{69.96±0.54} & \textbf{67.55±2.19} & \textbf{71.91±0.35} & \textbf{3.49±0.39}  \\
\hline
\end{tabular}
\end{table}

\begin{figure}[H]
\centering
\subfigure[Split SVHN]{\label{fig:subfig:a}
\includegraphics[width=0.35\linewidth]{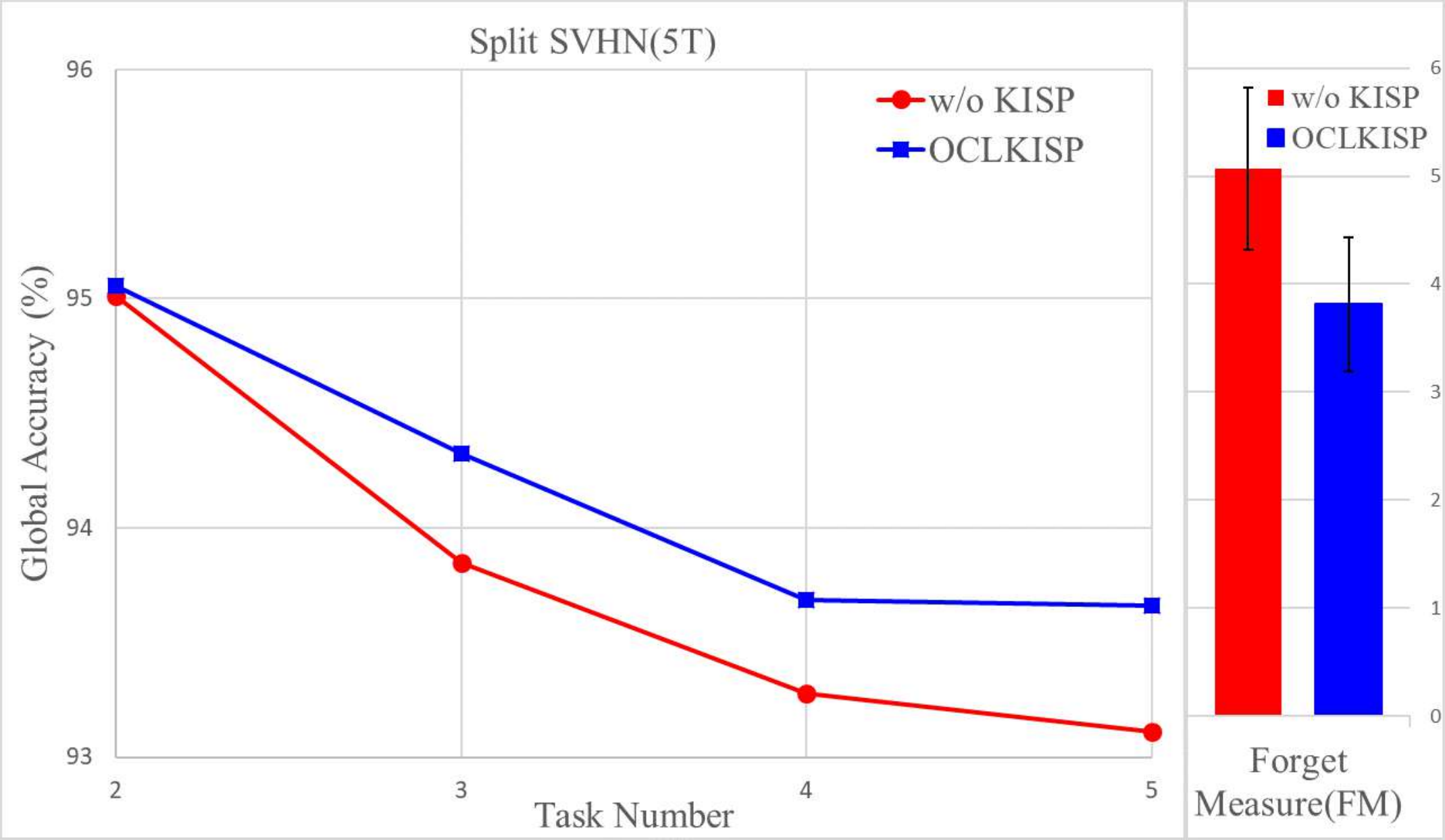}}
\hspace{0.01\linewidth}
\subfigure[Split CUB200]{\label{fig:subfig:b}
\includegraphics[width=0.35\linewidth]{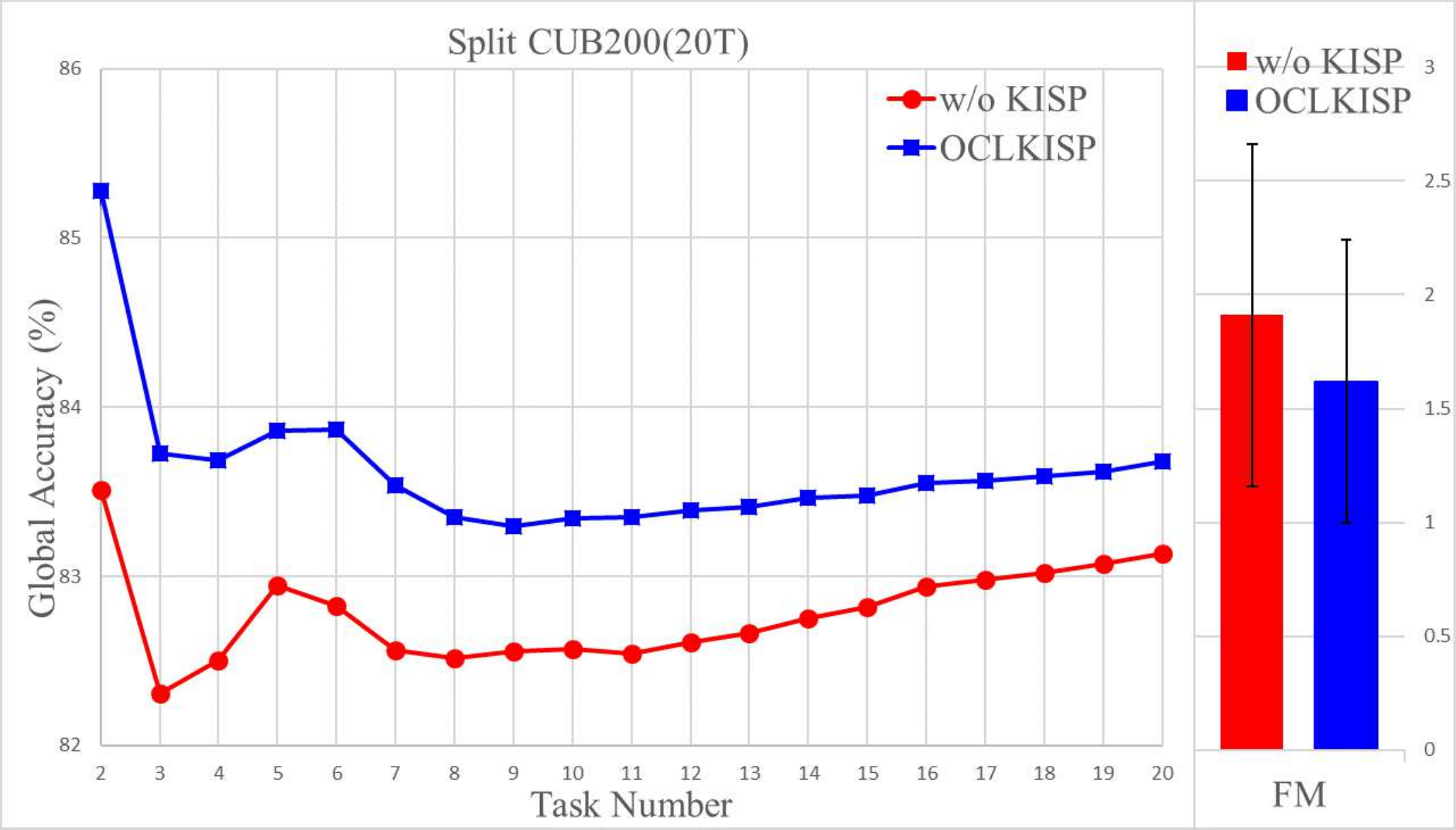}}
\vfill
\subfigure[Split Cifar100]{\label{fig:subfig:a}
\includegraphics[width=0.35\linewidth]{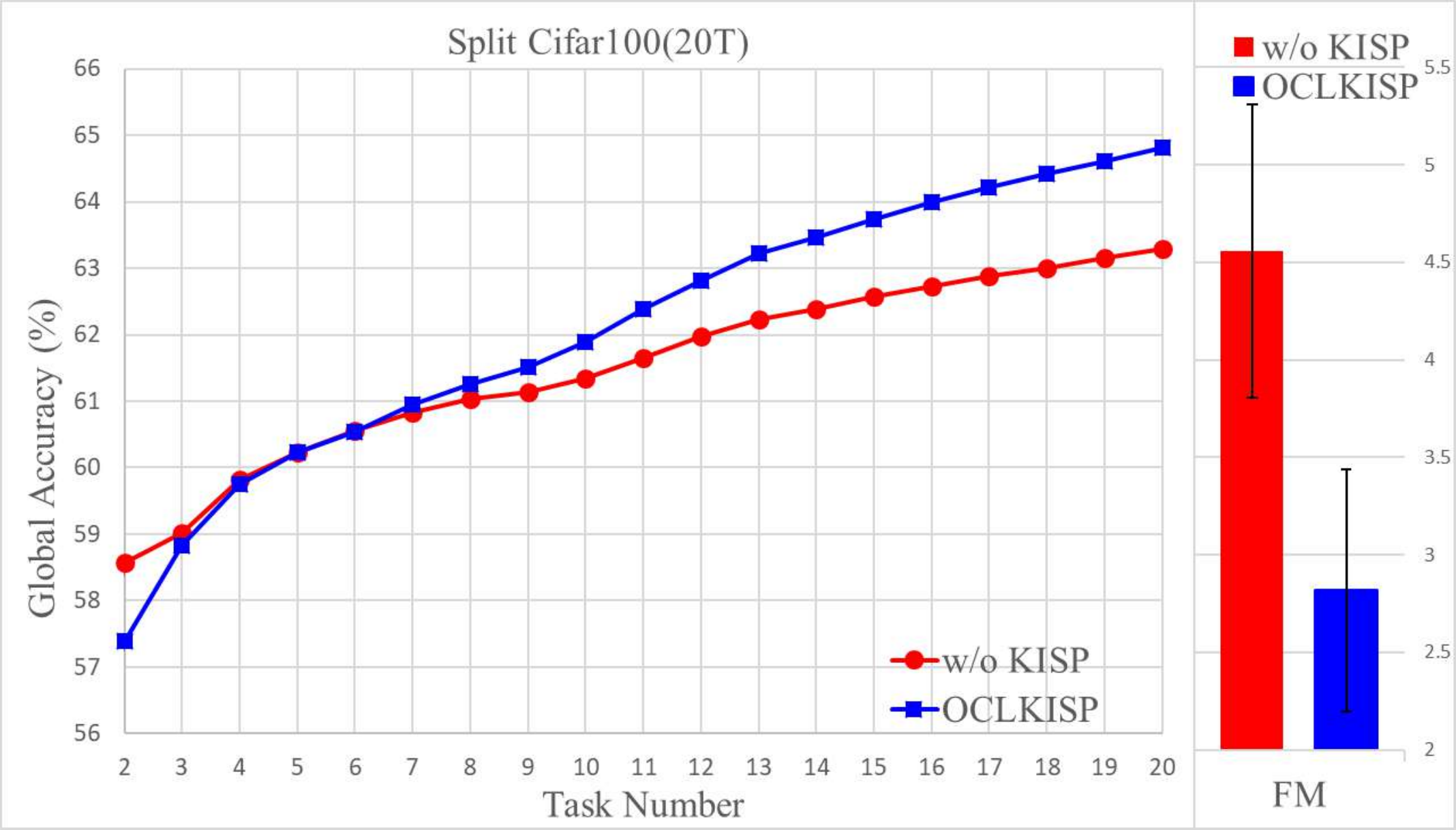}}
\hspace{0.01\linewidth}
\subfigure[Split Tiny-ImageNet]{\label{fig:subfig:b}
\includegraphics[width=0.35\linewidth]{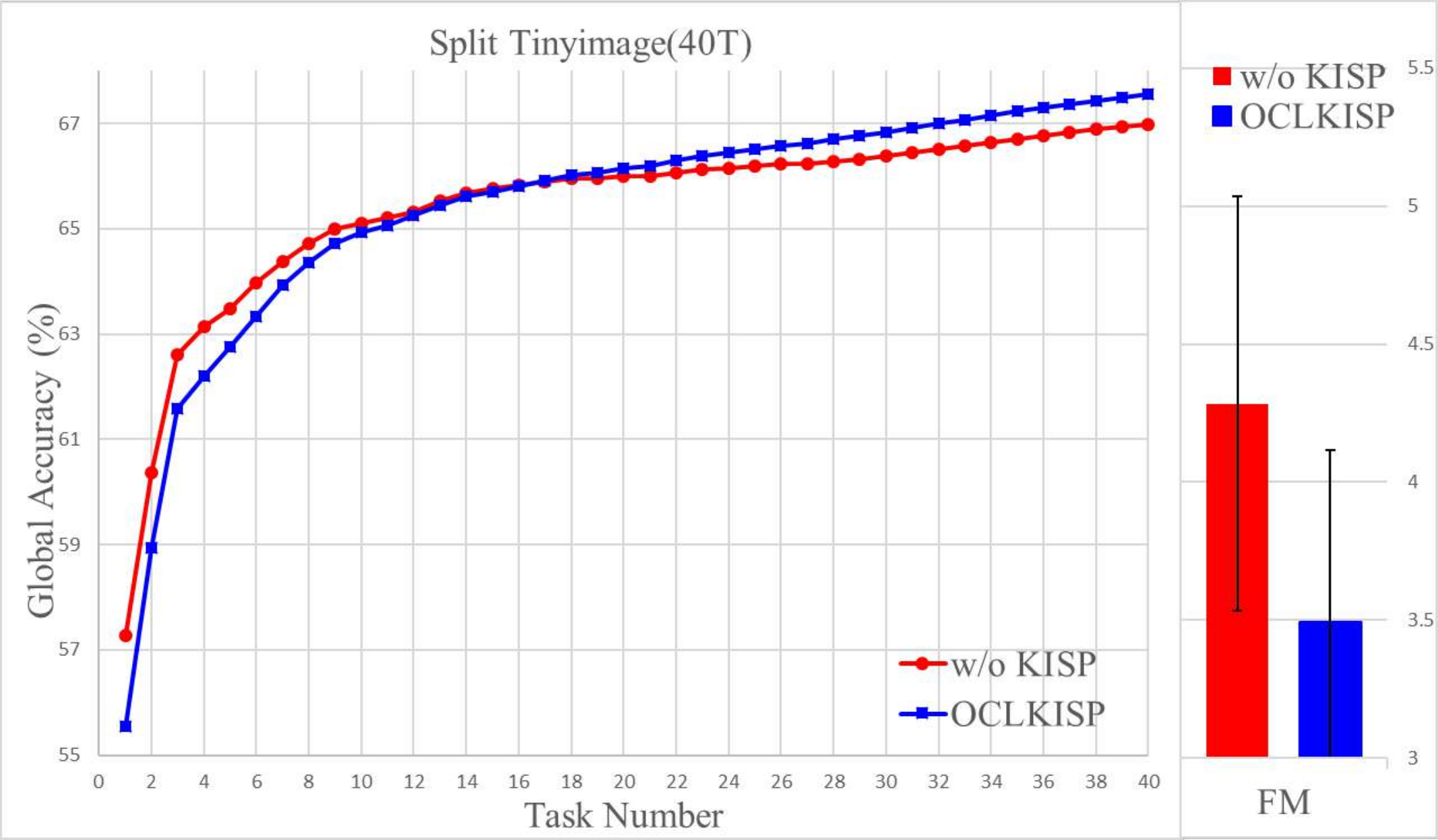}}
\caption{Performance variation of the global accuracy on various datasets}
\label{fig8}
\end{figure}

\subsubsection{Comparison of the effects of the different episodic memory sizes}

In this subsection, we study the effects of the different episodic memory sizes on different baselines. We evaluate five competing approaches GEM, ER, DER++, w/o KISP and OCLKISP on Split CIFAR and Split CUB datasets. The experimental results of every benchmark against the different episodic memory sizes are shown in Tables 4 and 5. Furthermore, Fig. 9 plots the final and global accuracy curves of each method. Firstly, we observe that all methods exhibit better overall performance when the episodic memory size increases. That means a larger memory is able to result in a better representation of the original training data, which can help remove catastrophic forgetting. Then, compared to w/o KISP, we found that the performance of OCLKISP consistently improves, e.g., offers 0.5\% to 2\% improvements on Split CIFAR100. These results illustrate that the idea of KISP indeed helps improve the performance of the model across different memory sizes. Moreover, the experimental results show that OCLKISP can outperform existing approaches considered in this work on almost all metrics across different memory sizes, which further validates the effectiveness of OCLKISP.

\begin{table}[H]
\centering
\caption{Numerical results for Split Cifar100 dataset with memory size M per task and 95\% confidence interval reported after symbol ±.}
\begin{tabular}{ccccccc} 
\hline
\multirow{2}{*}{Methods} & \multicolumn{3}{l}{M=25}             & \multicolumn{3}{l}{M=45}             \\
                         & FA         & GA         & FM         & FA         & GA         & FM         \\ 
\hline
GEM                      & 57.62±1.39 & 57.46±0.83 & 10.91±1.25 & 60.97±1.01 & 60.27±1.38 & 8.19±0.83  \\
ER                       & 57.78±1.65 & 58.12±0.63 & 12.65±1.65 & 61.83±1.17 & 60.05±1.21 & 8.19±0.96  \\
DER++                    & 61.93±1.30 & 58.64±1.76 & 7.37±0.92  & 66.39±0.87 & 62.05±3.09 & 4.90±0.84  \\
w/o KISP                 & 59.81±0.59 & 58.07±1.92 & 9.92±1.06  & 65.50±0.64 & 62.06±2.26 & 4.64±0.45  \\
OCLKISP                  & 60.65±1.48 & 58.42±1.55 & 8.93±1.42  & 66.32±0.49 & 62.52±2.53 & 4.14±0.24  \\ 
\hline
Methods                  & \multicolumn{3}{l}{M=65}             & \multicolumn{3}{l}{M=85}             \\ 
\hline
GEM                      & 62.63±0.58 & 61.18±1.04 & 7.08±0.41  & 64.24±1.03 & 62.30±2.20 & 5.29±0.82  \\
ER                       & 63.04±1.56 & 62.29±0.76 & 5.83±2.17  & 65.28±0.32 & 63.62±0.82 & 5.94±0.37  \\
DER++                    & 67.84±1.14 & 64.59±3.07 & 3.47±0.46  & 68.53±0.94 & 65.41±3.75 & 3.06±0.50  \\
w/o KISP                 & 66.03±1.34 & 63.30±1.70 & 4.56±0.94  & 68.83±0.59 & 65.81±2.65 & 3.21±0.38  \\
OCLKISP                  & 68.70±1.00 & 64.81±2.23 & 2.81±0.42  & 69.38±0.49 & 66.36±2.24 & 2.44±0.31  \\
\hline
\end{tabular}
\end{table}

\begin{table}[H]
\centering
\caption{Numerical results for Split CUB dataset with memory size M per task and 95\% confidence interval reported after symbol ±.}
\begin{tabular}{lllllll} 
\hline
\multirow{2}{*}{Methods} & \multicolumn{3}{l}{M=25}            & \multicolumn{3}{l}{M=45}             \\
                         & FA         & GA         & FM        & FA         & GA         & FM         \\ 
\hline
GEM                      & 74.40±2.03 & 75.59±0.94 & 7.03±1.76 & 80.27±0.80 & 78.54±0.59 & 2.88±1.00  \\
ER                       & 76.31±1.36 & 75.95±0.39 & 7.03±1.12 & 80.26±1.30 & 78.84±2.38 & 3.09±0.97  \\
DER++                    & 78.45±1.04 & 76.63±1.33 & 3.48±1.11 & 81.56±1.68 & 79.83±0.83 & 2.18±0.75  \\
w/o KISP                 & 79.73±0.76 & 80.17±0.86 & 4.97±0.54 & 83.70±0.22 & 82.31±1.00 & 1.84±0.23  \\
OCLKISP                  & 80.90±0.67 & 81.05±0.74 & 4.42±0.69 & 83.82±0.61 & 82.59±0.32 & 2.39±0.61  \\ 
\hline
Methods                  & \multicolumn{3}{l}{M=65}            & \multicolumn{3}{l}{M=85}             \\ 
\hline
GEM                      & 81.69±1.57 & 79.39±1.68 & 2.23±1.57 & 82.18±2.45 & 81.61±1.73 & 2.70±2.09  \\
ER                       & 81.42±1.07 & 81.05±0.51 & 3.59±0.93 & 82.35±1.11 & 81.67±0.89 & 2.21±1.45  \\
DER++                    & 83.32±0.46 & 81.02±0.51 & 1.22±0.56 & 83.91±0.59 & 82.16±2.57 & 1.15±0.76  \\
w/o KISP                 & 84.31±0.48 & 83.14±0.24 & 1.91±0.59 & 85.49±0.25 & 84.63±0.25 & 1.67±0.57  \\
OCLKISP                  & 85.08±0.51 & 83.45±0.73 & 1.62±0.49 & 85.72±0.85 & 84.84±0.40 & 1.79±0.03  \\
\hline
\end{tabular}
\end{table}

\begin{figure}[H]
\centering
\subfigure[Split SVHN]{\label{fig:subfig:a}
\includegraphics[width=0.35\linewidth]{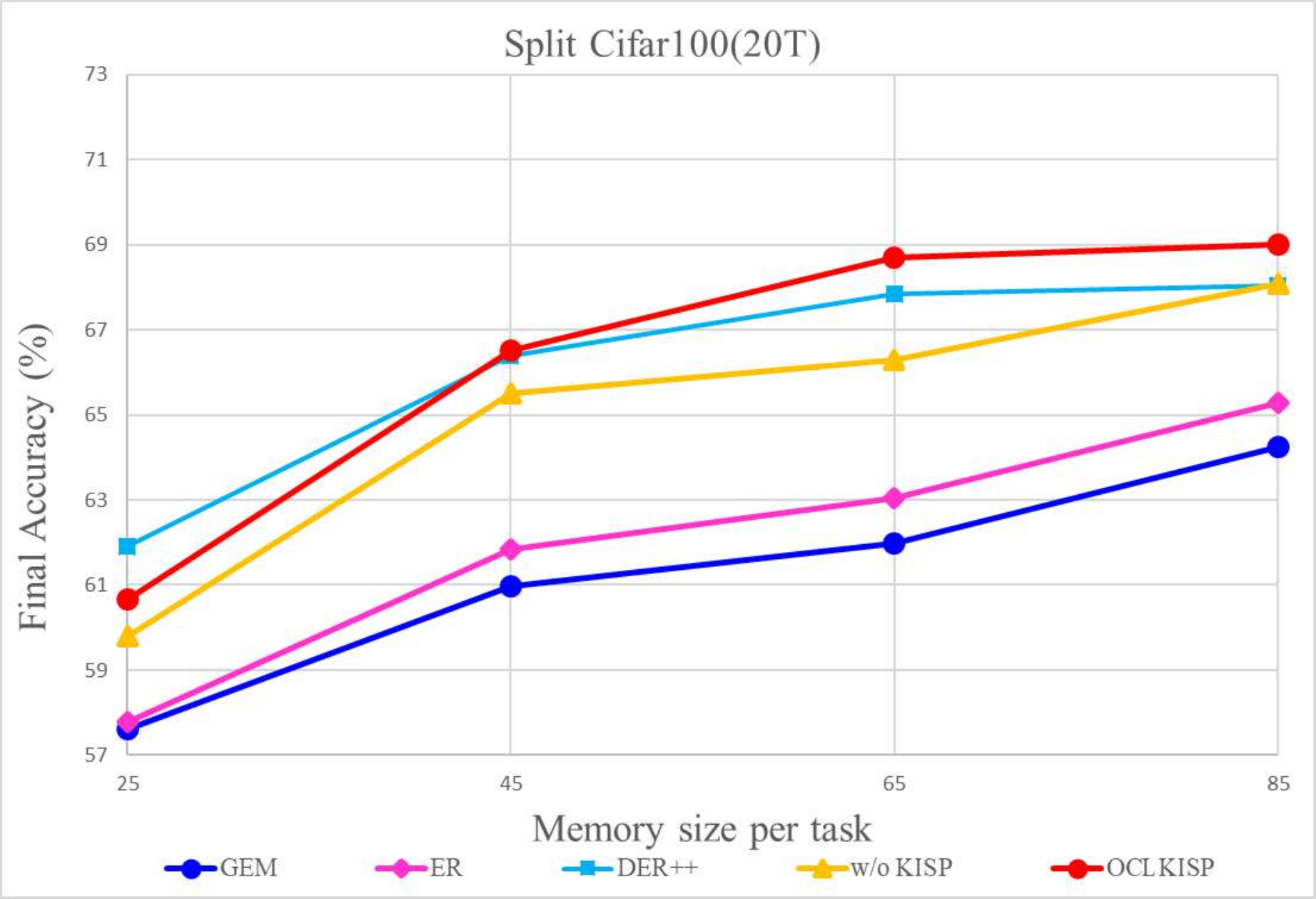}}
\hspace{0.01\linewidth}
\subfigure[Split CUB200]{\label{fig:subfig:b}
\includegraphics[width=0.35\linewidth]{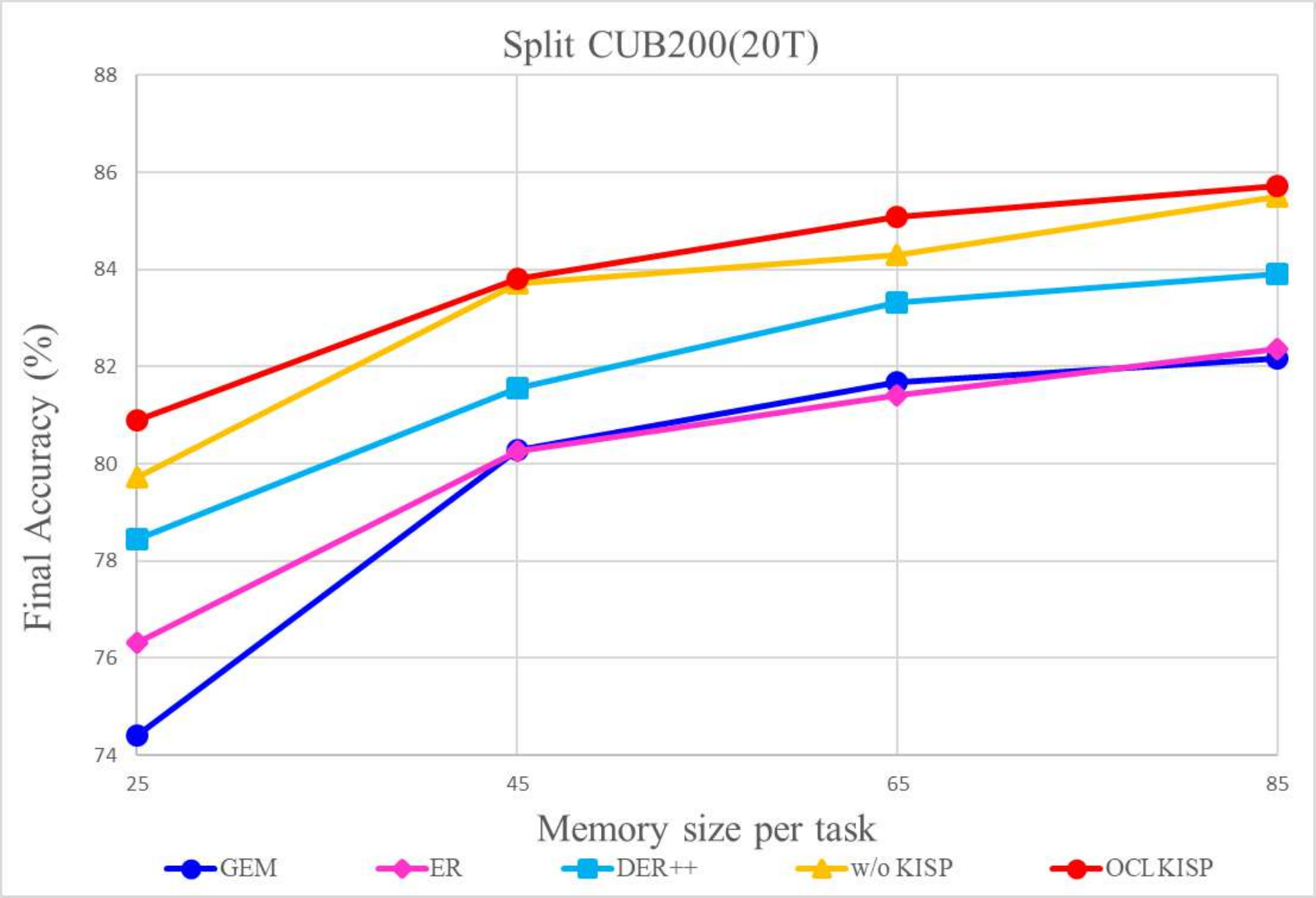}}
\vfill
\subfigure[Split Cifar100]{\label{fig:subfig:a}
\includegraphics[width=0.35\linewidth]{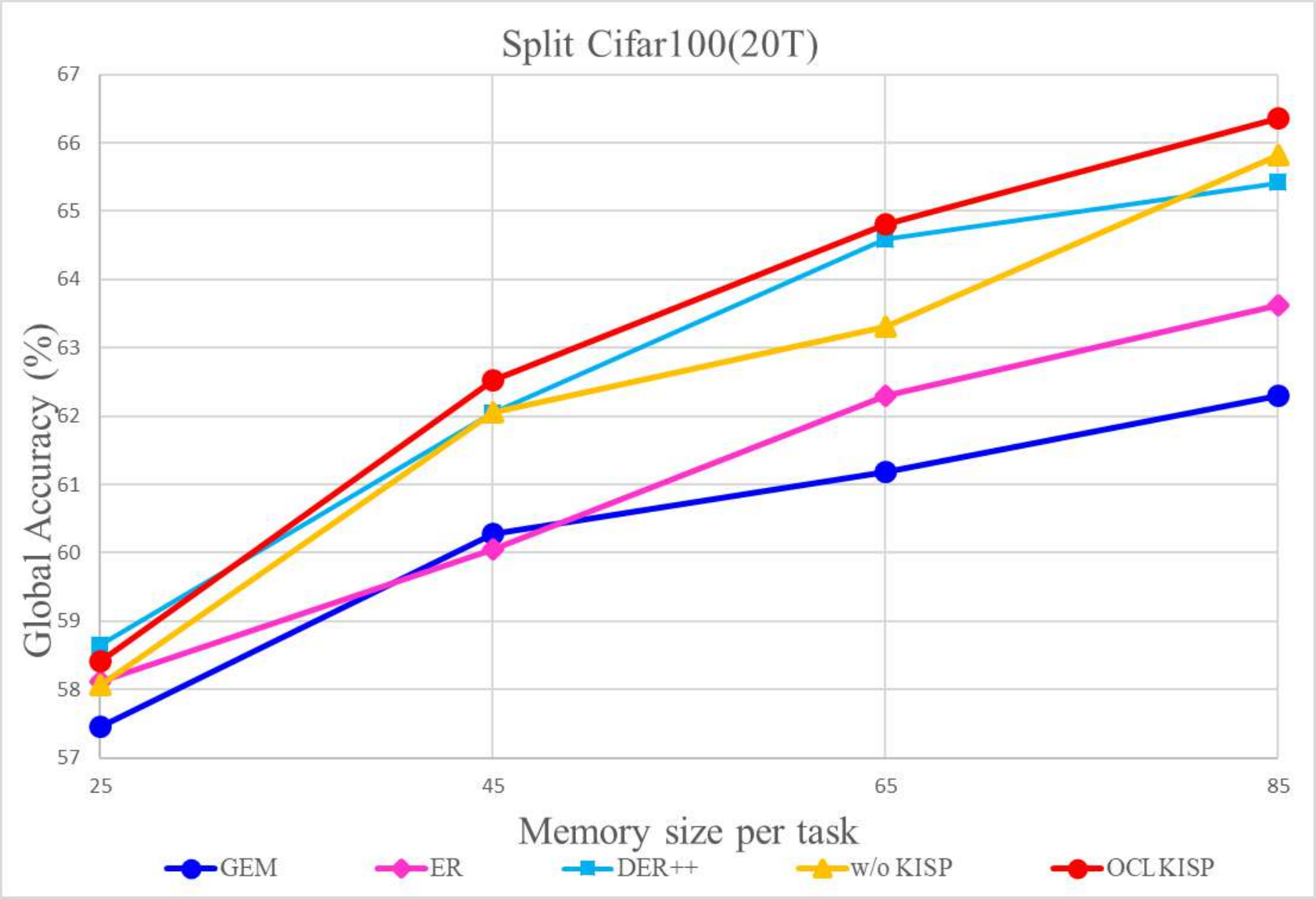}}
\hspace{0.01\linewidth}
\subfigure[Split Tiny-ImageNet]{\label{fig:subfig:b}
\includegraphics[width=0.35\linewidth]{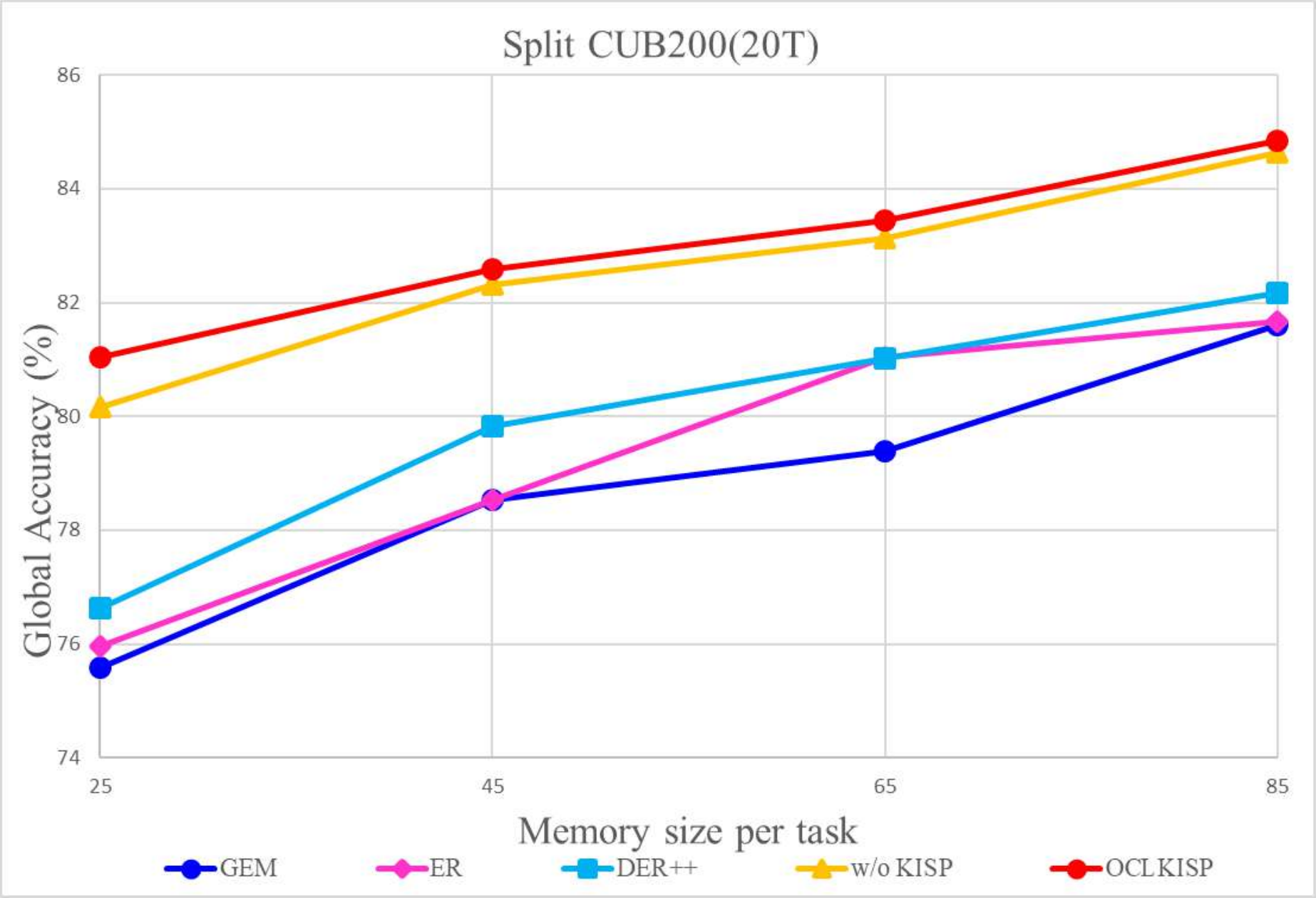}}
\caption{Comparison of the effects of the different episodic memory sizes on Split Cifar100 and Split CUB200 datasets. a) and b) plot the final accuracies with different memory sizes. c) and d) plot the global accuracies with different memory sizes.}
\label{fig9}
\end{figure}

\section{Conclusions and future work}

Continual learning aims to prevent catastrophic forgetting when new tasks are added. In this paper, we have studied the potentials and limitations of current popular continual learning paradigms. The experimental results illustrate the effectiveness of some state-of-the-art approaches, especially memory-based approaches. However, the learning bias between the current and previous tasks during experience replay is also a key problem, which can enhance forgetting. Furthermore, existing methods also neglect the transfer of structural knowledge in episodic memory which would be harmful to learning tasks sequentially. Inspired by this, this work introduces a novel continual learning method called Online Continual Learning via the Knowledge Invariant and Spread-out Properties (OCLKISP). The proposed approach can further transfer the structural characteristics of instances via knowledge invariant and spread-out properties when continually learning a sequence of tasks, which can also be beneficial to overcome forgetting caused by the learning bias. Finally, comprehensive experiments are conducted on various benchmark datasets to validate the effectiveness of our proposed paradigm.

In the future, there are several directions worth considering. Firstly, in this paper, we have illustrated the feasibility of our OCLKISP approach in classification settings and we can consider further extending our proposed method to more avenues of computer vision, such as continual face recognition and continual semantic segmentation. Besides, we can also consider some more realistic problems during the practical application, such as concept drift and abnormal detection. Furthermore, note that experience replay is an important means to overcome catastrophic forgetting, thus it would be interesting to study how to select the more representative samples which can better preserve latent decision boundaries for previously observed tasks. Another interesting extension to this work would be to use generative models to generate new examples which are used instead of episodic memory.

\bibliography{bibfile}

\end{document}